\documentclass[runningheads]{llncs}
\usepackage{graphicx}

\usepackage{tikz}
\usepackage{comment}
\usepackage{amsmath,amssymb} %
\usepackage{color}

\usepackage[accsupp]{axessibility}  %

\usepackage[pagebackref,breaklinks,colorlinks,breaklinks=true]{hyperref}
\usepackage{breakcites}

\usepackage{epsfig}
\usepackage{graphicx}
\usepackage{amsmath}
\usepackage{amssymb}

\usepackage{bbm}
\usepackage{inconsolata}
\usepackage{subcaption}
\usepackage{multirow}
\usepackage{booktabs}   %
\usepackage{makecell}   %
\usepackage{changepage} %
\usepackage{comment}

\usepackage[font=small]{caption}

\usepackage{wrapfig}

\usepackage{setspace}

\usepackage{colortbl}

\usepackage{pifont}
\newcommand{\cmark}{\ding{51}}

\newcommand{\cY}{{\cal Y}}
\newcommand{\cA}{{\cal A}}
\newcommand{\cO}{{\cal O}}
\newcommand{\bt}{{\bf t}}
\newcommand{\cT}{{\cal T}_\text{init}}
\newcommand{\cTfinal}{{\cal T}}

\newcommand{\cue}[1]{\texttt{C}_#1}

\usepackage{tabularx}
\newcolumntype{R}{>{\raggedleft\arraybackslash}X}

\usepackage{longtable}

\usepackage{inconsolata}

\newcommand{\figref}[1]{Fig\onedot~\ref{#1}}
\newcommand{\equref}[1]{Eq\onedot~\eqref{#1}}
\newcommand{\secref}[1]{Sec\onedot~\ref{#1}}
\newcommand{\tabref}[1]{Tab\onedot~\ref{#1}}

\usepackage[normalem]{ulem}

\newcommand{\argmax}{\mathop{\mathrm{argmax}}}

\usepackage{adjustbox}

\makeatletter
\usepackage{xspace}
\def\@onedot{\ifx\@let@token.\else.\null\fi\xspace}
\DeclareRobustCommand\onedot{\futurelet\@let@token\@onedot}

\def\eg{\emph{e.g}\onedot} 
\def\ie{\emph{i.e}\onedot} 
 
 \def\vs{\emph{vs}\onedot}
 
\def\etal{\emph{et al}\onedot}

\usepackage{color, colortbl}
\definecolor{Gray}{gray}{0.9}
\newcolumntype{g}{>{\columncolor{Gray}}r}
\definecolor{highlightRowColor}{rgb}{0.95, 0.95, 1}

\newcommand*{\belowrulesepcolor}[1]{%
  \noalign{%
    \kern-\belowrulesep 
    \begingroup 
      \color{#1}%
      \hrule height\belowrulesep 
    \endgroup 
  }%
} 
\newcommand*{\aboverulesepcolor}[1]{%
  \noalign{%
    \begingroup 
      \color{#1}%
      \hrule height\aboverulesep 
    \endgroup 
    \kern-\aboverulesep 
  }%
} 

\newcommand{\beginsupplement}{
    \setcounter{table}{0}
    \renewcommand{\thetable}{S\arabic{table}}%
    \setcounter{figure}{0}
    \renewcommand{\thefigure}{S\arabic{figure}}%
    \setcounter{equation}{0}
    \renewcommand{\theequation}{S\arabic{equation}}
}

\usepackage[resetlabels]{multibib}
\newcites{supp}{References}

\makeatletter
\g@addto@macro\normalsize{%
  \setlength\abovedisplayskip{4pt}
  \setlength\belowdisplayskip{5pt}
}
\makeatother

\newcommand{\csection}[1]{
    \section{#1}
}

\newcommand{\csubsection}[1]{
    \subsection{#1}
}

\usepackage{contour}
\usepackage[normalem]{ulem}

\contourlength{0.8pt}

\newcommand{\myuline}[1]{%
  \uline{\phantom{#1}}%
  \llap{\contour{white}{#1}}%
}

\usepackage[capitalize]{cleveref}
\crefname{section}{Sec.}{Secs.}
\Crefname{section}{Section}{Sections}
\Crefname{table}{Table}{Tables}
\crefname{table}{Tab.}{Tabs.}
\crefformat{footnote}{#2\footnotemark[#1]#3}

\begin{document}
\pagestyle{headings}
\mainmatter
\def\ECCVSubNumber{4622}  %

\title{Initialization and Alignment \\for Adversarial Texture Optimization} %

\titlerunning{Initialization and Alignment for AdvTex}
\authorrunning{X. Zhao, Z. Zhao, A. G. Schwing.}

\author{Xiaoming Zhao, Zhizhen Zhao, Alexander G. Schwing}
\institute{
University of Illinois, Urbana-Champaign \\
\email{\{xz23, zhizhenz, aschwing\}@illinois.edu} \\
\url{https://xiaoming-zhao.github.io/projects/advtex_init_align}
}

\maketitle

\begin{abstract}
While  recovery of geometry from image and video data has received a lot of attention in computer vision, methods to capture the texture for a given geometry are less mature. 
Specifically, classical methods for texture generation  often assume  clean geometry and reasonably well-aligned image data. While very recent methods, \eg, adversarial texture optimization, better handle lower-quality data obtained from hand-held devices, we find them to still struggle frequently. 
To improve robustness, particularly of recent adversarial texture optimization, we develop an explicit initialization and an alignment procedure. It deals with complex geometry due to a robust mapping of the geometry to the texture map and a hard-assignment-based initialization. It  deals with misalignment of geometry and images by integrating fast image-alignment into the texture refinement optimization. %
We demonstrate  efficacy of our texture generation on a dataset of 11 scenes with a total of 2807 frames, observing 7.8\% and 11.1\% relative improvements regarding perceptual and sharpness measurements.

\end{abstract}

\csection{Introduction}
Accurate scene reconstruction is one of the major goals in computer vision. Decades of research have been devoted  to developing robust methods like `Structure from Motion,' `Bundle Adjustment,' and more recently also single view reconstruction techniques. 
While reconstruction of geometry from image and video data has become increasingly popular and accurate in recent years, recovered 3D models remain often pale because  textures aren't considered.

Given a reconstructed 3D model of a scene consisting of triangular faces, and given a sequence of images depicting the scene, texture mapping aims to find for each triangle a suitable texture. 
The problem of automatic texture mapping has been  studied  in different areas since    late 1990 and early 2000. 
For instance, in the graphics community~\cite{Debevec1996,Marshner1998,Neugebauer1999}, in computer vision~\cite{ThierryCVPR2007,Lempitsky2007,Sinha2008Interactive3A}, architecture~\cite{Grammatikopoulos2007} and photogrammetry~\cite{ElHakim2003}.
Many of the proposed algorithms work very well in a controlled lab-setting where geometry is known perfectly, or in a setting where accurate 3D models are available from a  3D laser scanner.

However, applying  texture mapping techniques to noisy mesh geometry obtained \emph{on the fly} from a recent LiDAR equipped iPad reveals missing robustness because of multiple reasons: 1) images and 3D models are often not perfectly aligned; 2) 3D models are not accurate and the obtained meshes aren't necessarily manifold-connected.
Even recent techniques for mesh flattening~\cite{Sawhney2018BoundaryFF} and texturing~\cite{Fu2018TextureMF,Zhou2014ColorMO,Huang2020AdversarialTO} result in surprising artifacts due to streamed geometry and pose inaccuracies as shown later.

\begin{figure}[t]
    \centering
    \includegraphics[width=0.5\linewidth]{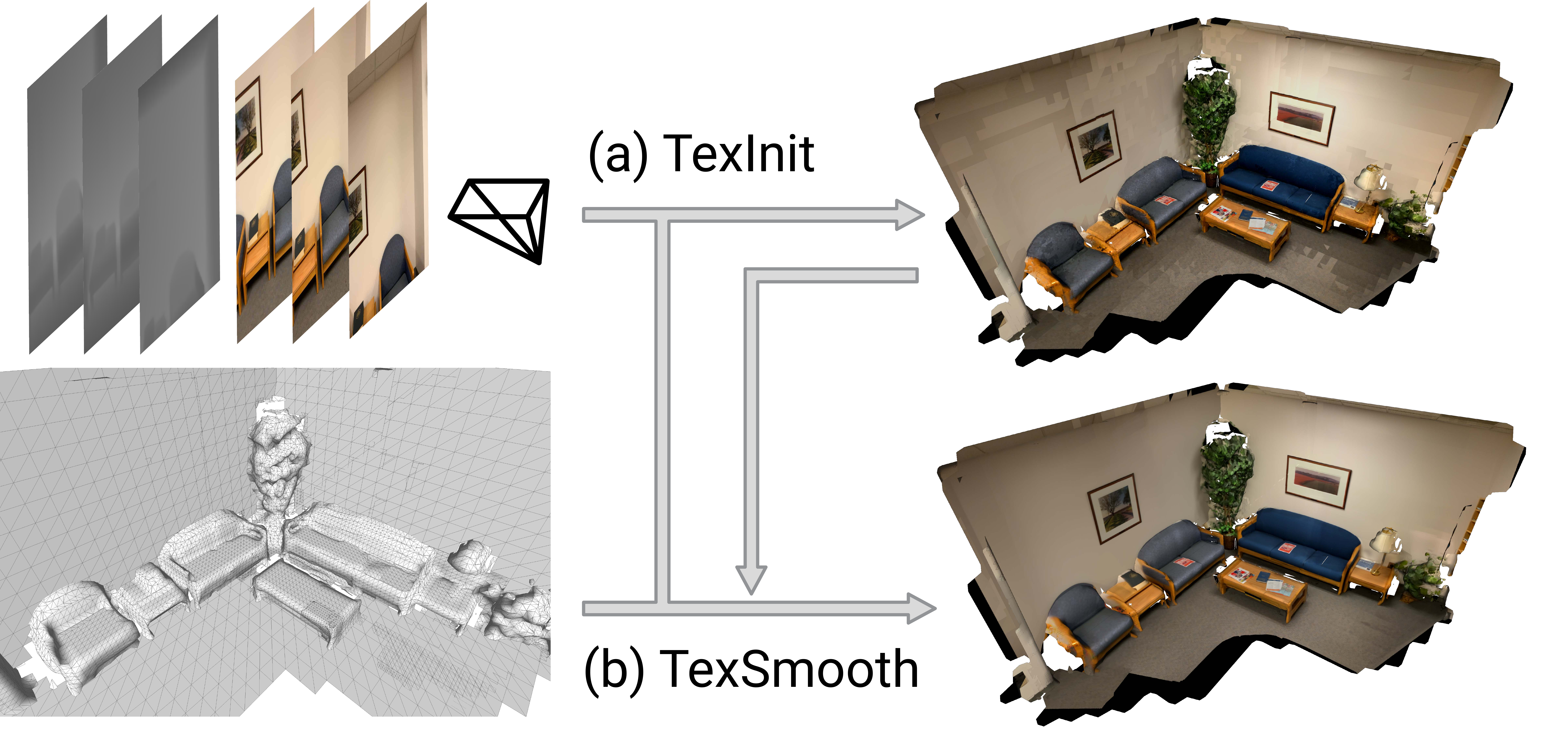}
    \caption{
    We study texture generation given RGBD images with associated camera parameters as well as a reconstructed mesh.
    \textbf{(a) \texttt{TexInit}}: we initialize the texture using an assignment-based texture generation framework.
    \textbf{(b) \texttt{TexSmooth}}: a data-driven adversarial loss is utilized to optimize out artifacts incurred in the assignment step.
    }
    \label{fig:teaser}
\end{figure}

To address this robustness issue we find equipping of the recently-proposed adversarial texture optimization technique~\cite{Huang2020AdversarialTO} with classical initialization and alignment techniques to be remarkably effective.
Without the added initialization and alignment, we find current methods don't produce high-quality textures.  
Concretely and as illustrated in \figref{fig:teaser},  we aim for texture generation which operates on a sequence of images and corresponding depth maps as well as their  camera parameters. Moreover, we assume the 3D model to be given and fixed. 
Importantly, we consider a streaming setup, with all data  obtained \emph{on the fly}, and not further processed,~\eg,~via batch structure-from-motion.
The setup is ubiquitous and the form of data can be acquired easily from consumer-grade devices these days, \eg, from a recent iPad or iPhone~\cite{arkit2021}. 
We aim to translate this  data into texture maps.
For this, we first flatten the triangle mesh using recent advances~\cite{Sawhney2018BoundaryFF}. We then use a Markov Random Field (MRF)  to resolve  overlaps in flattened meshes for non-manifold-connected data. In a next step we determine the image frame from which to extract the texture of each mesh triangle using a simple optimization. %
We refer to this as \texttt{TexInit}, which permits to obtain a high-quality initialization for subsequent refinement.
Next we address  inaccuracies in camera poses and in geometry by automatically shifting images using the result of a fast Fourier transformation (FFT)~\cite{AnutaGeo1970}.
The final optimized texture is obtained by integrating this FFT-alignment component into adversarial optimization~\cite{Huang2020AdversarialTO}. 
We dub this stage \texttt{TexSmooth}.
The obtained texture can be used in any 3D engine for downstream applications.

To study efficacy of the proposed framework we acquire 11 complex scenes using a recent iPad.
We establish  accuracy of the proposed technique to generate and use the texture by showing that the quality of rendered views is superior to prior approaches on these scenes. 
Quantitatively, our framework improves prior work by 7.8\% and 11.1\% relatively with respect to perceptual (LPIPS~\cite{Zhang2018TheUE}) and sharpness ($S_3$~\cite{Vu2012bfSA}) measurements respectively. Besides, our framework improves over baselines on ScanNet~\cite{Dai2017ScanNetR3}, demonstrating the ability to generalize.

\csection{Related Work}
We aim for accurate recovery of texture  for a reconstructed 3D scene from a sequence of  RGBD images.
For this, a variety of techniques have been proposed, which can be roughly categorized into four groups: 1) averaging-based;  2) warping-based; 3) learning-based; and 4)  assignment-based. Averaging-based methods find all views within which a point is visible and combine the color observations. Warping-based approaches either distort or synthesize source images to handle mesh misalignment or camera drift. Learning-based ones learn the texture representation. Assignment-based methods attempt to find the best view and `copy' the observation into a texture. 
We review these groups next:

\noindent\textbf{Averaging-based:} 
Very early work by Marshner~\cite{Marshner1998} estimates the parameters of a bidirectional reflectance distribution function (BRDF) for every point on the texture map. To compute this estimation, all observations from the recorded images where the point is visible are used. Similar techniques have been investigated in subsequent work~\cite{Bernardini2001}.

Similarly, to compute a texture map,
\cite{Neugebauer1999} and~\cite{Debevec1996} perform a weighted blending of all recorded images. The weights take visibility and other factors into account. The developed approaches are semi-automatic as they require  an initial estimate of the camera orientations which is obtained from interactively selected point correspondences or marked lines.
Multi-resolution textures~\cite{Ofek1997}, face textures~\cite{Pighin1998} and blending~\cite{Niem1999,BaumbergBMVC2002} have also been studied. %

\noindent\textbf{Warping-based:} 
Aganj~\etal~\cite{Aganj2009MultiviewTO} morph each source image to align to the mesh.
Furthermore,~\cite{Zhou2014ColorMO,Huang20173DliteTC} propose to optimize camera poses
and image warping parameters jointly. However, this line of vertex-based optimization has stringent requirements on the mesh density and cannot be applied to a sparse mesh.
More recently, Bi~\etal~\cite{Bi2017PatchbasedOF} follow patch-synthesis~\cite{Wexler2004SpacetimeVC,Barnes2009PatchMatchAR} to re-synthesize aligned target images for each source view. However, such methods require costly multiscale optimization to avoid a large number of local optima.  In contrast, the proposed approach does not require those techniques.  

\noindent\textbf{Learning-based:} 
Recently, learning-based methods have been introduced for texture optimization.
Some works focus on specific object and scene  categories~\cite{Goel2020ShapeAV,Saito2017PhotorealisticFT} while we do not  make such assumptions.
Moreover, learned representations,~\eg,~neural textures, have also been developed~\cite{Thies2019DeferredNR,Sitzmann2019DeepVoxelsLP,Aliev2020NeuralPG}.
Meanwhile, generative models are developed to synthesize a holistic texture from a single image or pattern~\cite{Henzler2020LearningAN,Oechsle2019TextureFL} while we focus on texture reconstruction.
AtlasNet~\cite{Groueix2018AtlasNetAP} and NeuTex~\cite{Xiang2021NeuTexNT} focus on learning a 3D-to-2D mapping, which can be utilized in texture editing, while we focus on reconstructing realistic textures from source images.
The recently-proposed adversarial texture optimization~\cite{Huang2020AdversarialTO} utilizes adversarial training to reconstruct the texture. However, despite advances, adversarial optimization still struggles  with   misalignments. We improve this shortcoming via an explicit high-quality initialization and an efficient  alignment module.

\noindent\textbf{Assignment-based:} 
Classical assignment-based methods operated within controlled environments~\cite{Rocchini1999,Rocchini2002,ElHakim2003,Frueh2004} or utilized special camera rigs~\cite{Frueh2004,Grammatikopoulos2007}. These  works suggest computing for each vertex a set of `valid' images,
which are subsequently refined by iterating over each vertex and adjusting the assignment to obtain more consistency. Finally,  texture data is `copied' from the images. 
In contrast, we aim to create a texture in an uncontrolled setting. %
Consequently, 3D geometry is not accurate and very noisy.
Other early work~\cite{Duan2003,Lensch2001,HEsteban2004,Wuhrer2006,Rocchini2002} focuses on closed surfaces and small-scale meshes, making them not applicable to our setting.
More recently, upon finding the best texture independently for each face using cues like visibility, orientation, resolution, and distortion, refinement techniques like texture coordinate optimization, color adjustments, or scores-based optimization have  been discussed~\cite{PanGM2015,Waechter2014LetTB,AbdelhafizJSS2020}. 

\begingroup
\setlength{\columnsep}{4pt}
\begin{wrapfigure}{r}{0.3\columnwidth}
\centering
  \begin{center}
    \includegraphics[width=0.3\columnwidth]{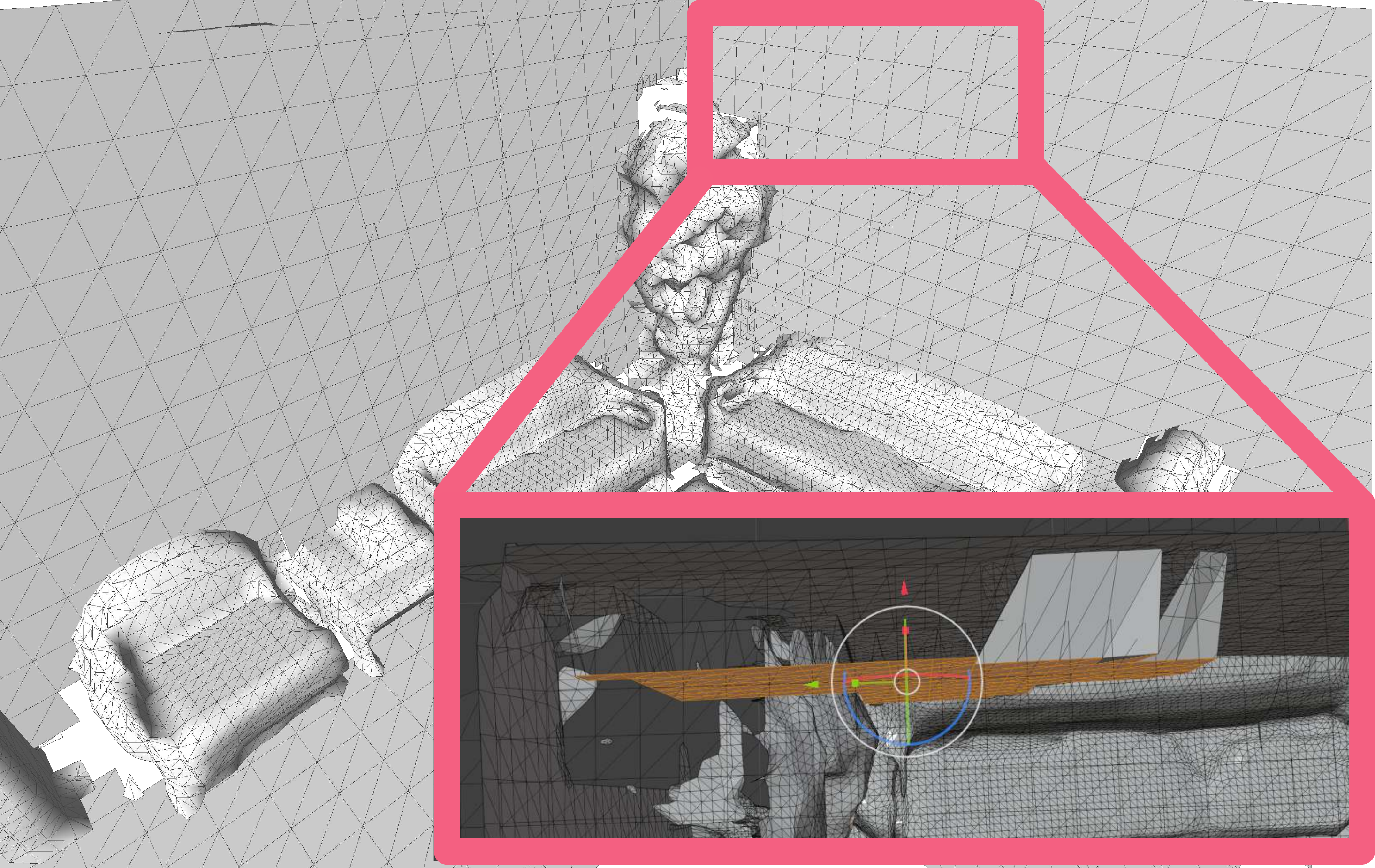}
  \end{center}
  \caption{Noisy geometry. The wall has two layers.}
  \label{fig: scene complexity}
\end{wrapfigure}
Related to our  approach are  methods that formulate texture selection using a Markov Random Field (MRF)~\cite{Lempitsky2007,Shu2016,Fu2018TextureMF}.
Shu \etal~\cite{Shu2016} suggest visibility as the data term and employ texture smoothness to reduce transitions.
Lempitsky \etal~\cite{Lempitsky2007} study color-continuity which integrates over  face seams. Fu \etal~\cite{Fu2018TextureMF} additionally use the projected 2D face area to select a texture assignment for each face. However, noisy geometry like the one shown in~\figref{fig: scene complexity}, makes it difficult for assignment-based methods to yield high quality results, which we will show later. 
Therefore, different from these methods, we address texture drift in a data-driven  refinement procedure rather than in an assignment stage. 

\endgroup

\csection{Approach}

We want to automatically create the texture $\cTfinal$ from a set of RGBD images $I=\{I_1, \dots, I_T\}$, for each of which we also know camera parameters $\{p_t\}_{t=1}^T$,~\ie, extrinsics and intrinsics. We are also given a triangular scene mesh $M = \left\{ \texttt{Tri}_i \right\}_{i=1}^{\vert M \vert}$, where $\texttt{Tri}_i$ denotes the $i$-th triangle. This form of data is easily accessible from commercially available consumer devices, \eg, a recent iPhone or iPad.

We construct the texture $\cTfinal$ in two steps that combine advantages of assignment-based and learning-based techniques:
1) \texttt{TexInit}: we generate a texture initialization $\cT\in\mathbb{R}^{H\times W\times 3}$ of height $H$, width $W$ and $3$ color channels in an assignment-based manner (\secref{sec: atlas gen});
2) \texttt{TexSmooth}: we then refine $\cT$ with an improved data-driven adversarial optimization that integrates an efficient alignment procedure (\secref{sec: tex smooth}). Formally, the final texture $\cTfinal$ is computed via 
\begin{align}
    &\cTfinal = \texttt{TexSmooth}\left( \cT, \{I_t\}_{t=1}^T, \{p_t\}_{t=1}^T, M \right), \nonumber \\
    \text{where } &\cT = \texttt{TexInit}\left( \{I_t\}_{t=1}^T, \{p_t\}_{t=1}^T, M \right).
\end{align}
We detail each component next.

\csubsection{Texture Initialization (\texttt{TexInit})}\label{sec: atlas gen}

\begin{figure*}[t]
    \centering
    \includegraphics[width=\textwidth]{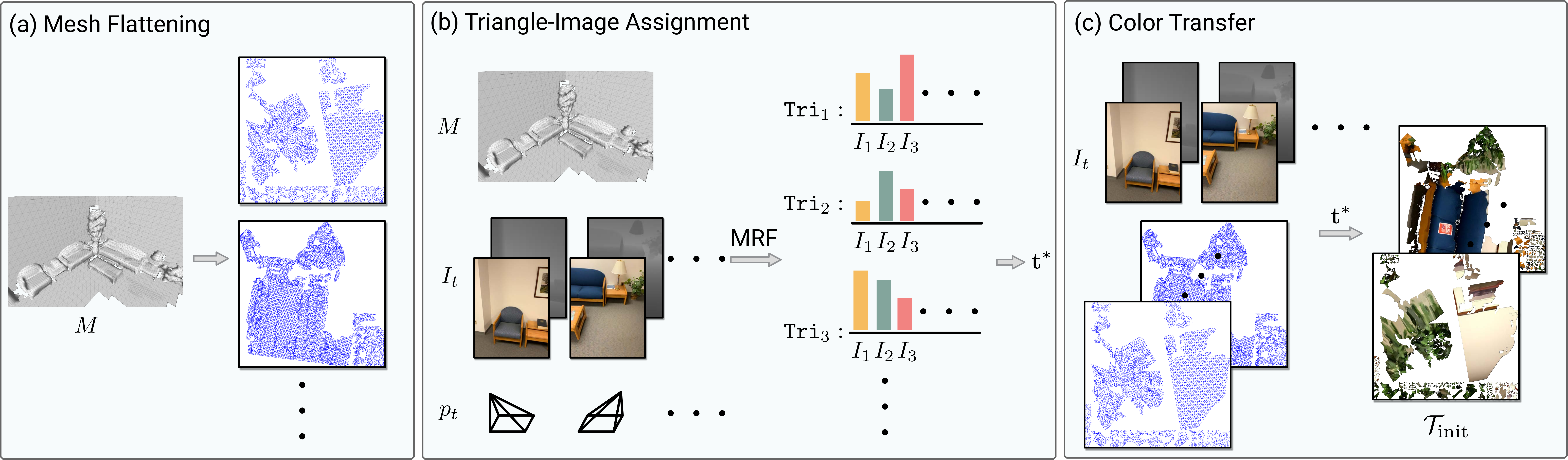}
    \caption{\textbf{Texture initialization \texttt{TexInit}} (\secref{sec: atlas gen}). \textbf{(a)} Mesh flattening: we flatten a 3D mesh into the 2D plane using overlap detection. \textbf{(b)} Triangle-image assignment: we develop a simple formulation to compute the triangle-image assignment $\bt^\ast$ from mesh $M$, frames $I_t$ and camera parameters $p_t$. We assign frames to each triangle $\texttt{Tri}_i$ based on  $\bt^\ast$. \textbf{(c)} Color transfer: based on the flattened mesh in \textbf{(a)} and the best assignment $\bt^\ast$ from \textbf{(b)}, we generate the texture $\cT$.}
    \label{fig:overview}
\end{figure*}

The proposed approach to obtain the texture initialization $\cT$ is outlined in \figref{fig:overview} and %
consists of following three steps: 
1) We flatten the provided mesh $M$.
For this we detect overlaps within the flattened mesh, which may happen due to the fact that we operate with  general meshes that are not guaranteed to have a manifold connectivity. Overlap detection ensures that every triangle is assigned a unique position in the texture. 
2) We identify for each triangle the `best' texture index $\bt^\ast$. Hereby, `best' is defined using cues like visibility and color consistency.
3) After identifying the index $\bt^\ast = (t_1^\ast, \dots, t_{|M|}^\ast)$ for each triangle, we create the texture $\cT$ by transferring for all $(u,v)\in[1, \dots, W]\times[1, \dots, H]$ locations in the texture, the  RGB data %
from the corresponding location $(a,b)$ in image $I_t$. %

\noindent\textbf{\myuline{1) Mesh Flattening:}} 
In a first step, as illustrated in \figref{fig:overview} (left), we flatten the given  mesh $M$. For this we use the recently proposed boundary first flattening (BFF) technique~\cite{Sawhney2018BoundaryFF}. The  flattening is fully automatic, with distortion mathematically guaranteed to be as low or lower than any other conformal mapping.

\begin{figure}[!t]
\begin{minipage}[t]{\columnwidth}
\centering
\begin{minipage}[t]{.35\columnwidth}
    \centering
    \begin{subfigure}{\textwidth}
    \centering
        \includegraphics[width=0.6\linewidth]{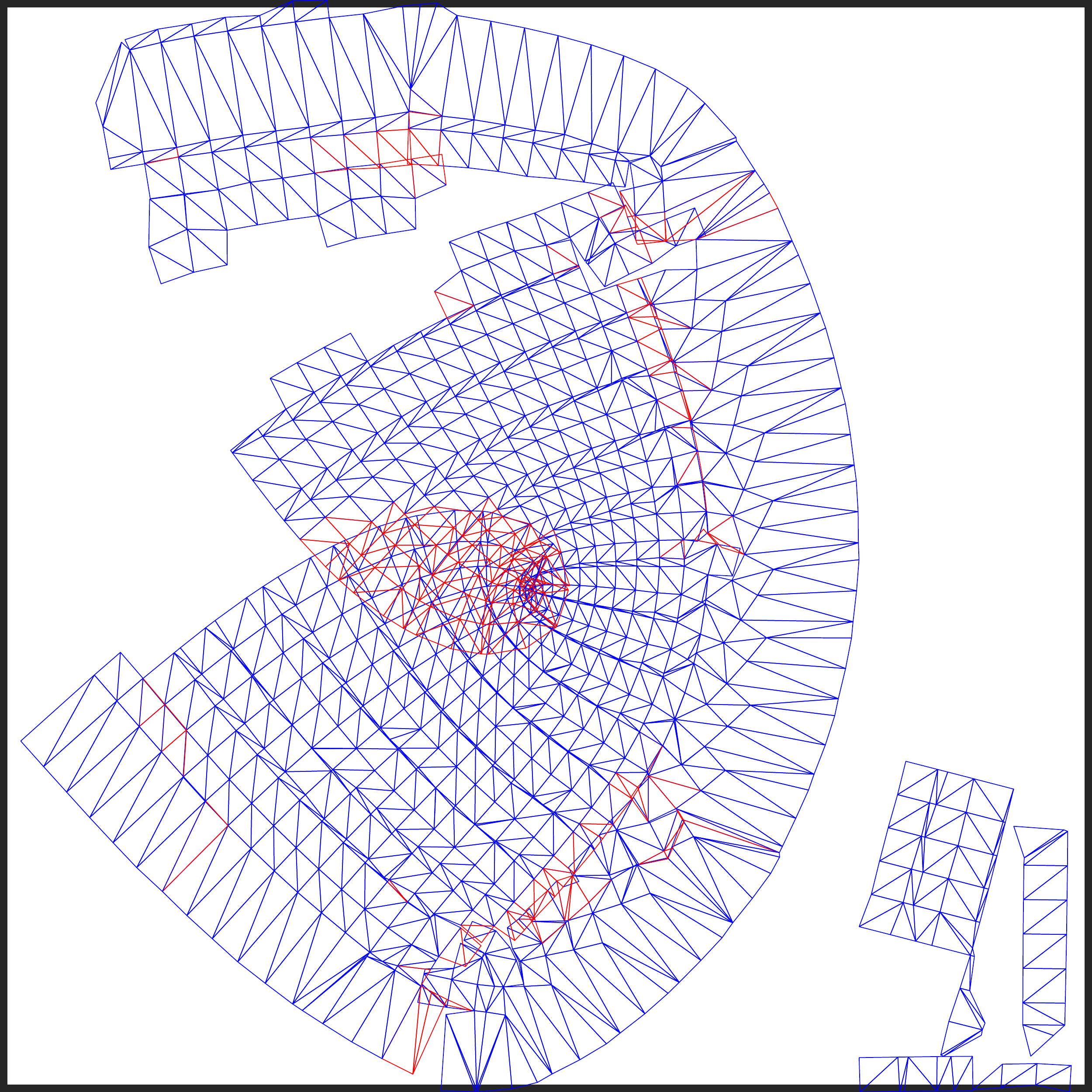}
        \captionsetup{width=\linewidth}
        \caption{Flattened mesh overlaps.}
        \label{fig: mesh flatten overlap}
    \end{subfigure}%
    \hspace{0.01\textwidth}
    \begin{subfigure}{\textwidth}
    \centering
        \includegraphics[width=0.6\linewidth]{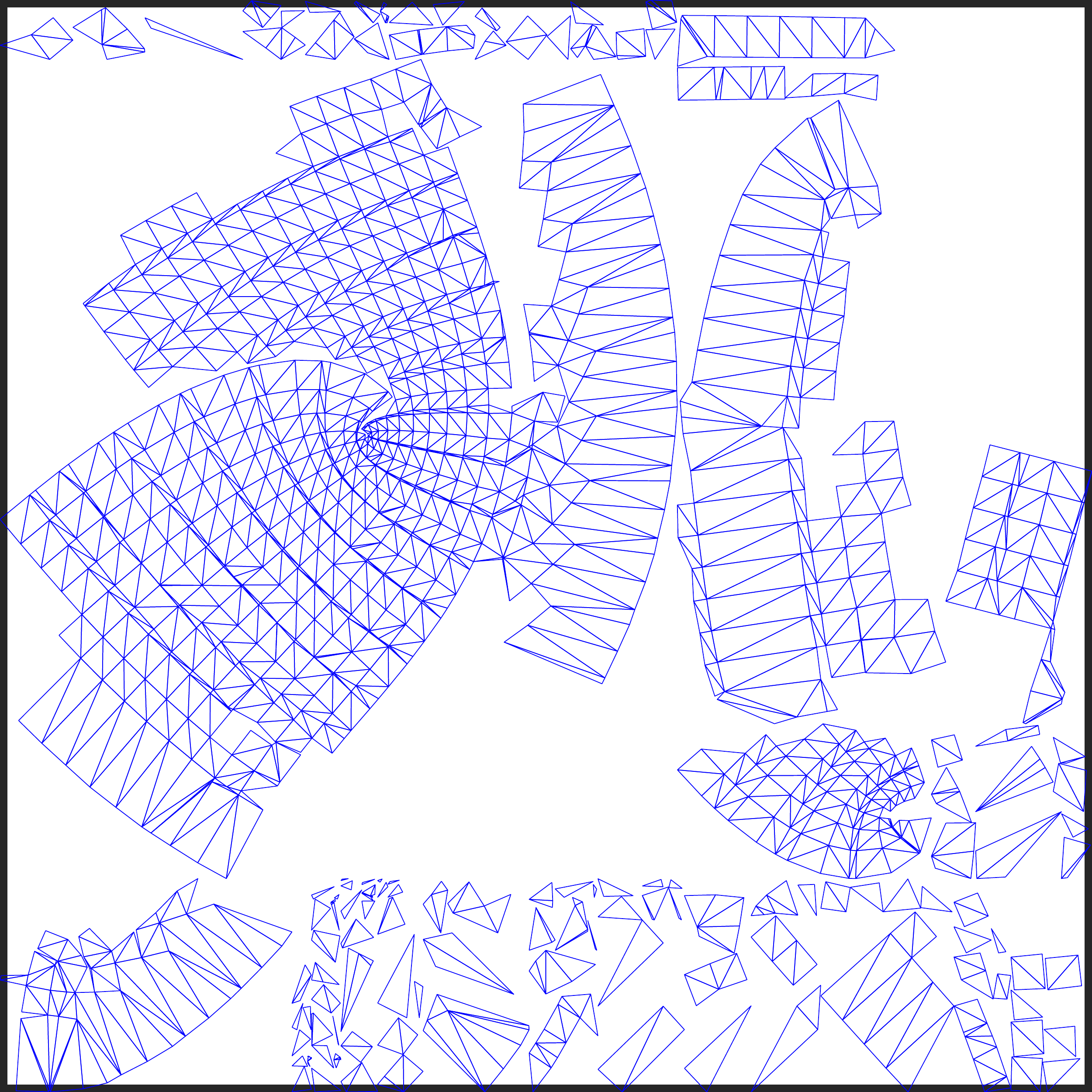}
        \captionsetup{width=\linewidth}
        \caption{Overlap-free.}
        \label{fig: mesh flatten non-overlap}
    \end{subfigure}%
    \caption{\textbf{Flattening.} \textbf{(a)} Red triangles indicate where overlap happens. \textbf{(b)} The proposed method (\secref{sec: atlas gen}) resolves this issue while keeping connectivity of areas.}
\label{fig:overlap}
\end{minipage}
\hfill
\begin{minipage}[t]{.62\columnwidth}
    \centering
    \begin{subfigure}{0.45\columnwidth}
    \centering
        \includegraphics[width=\linewidth]{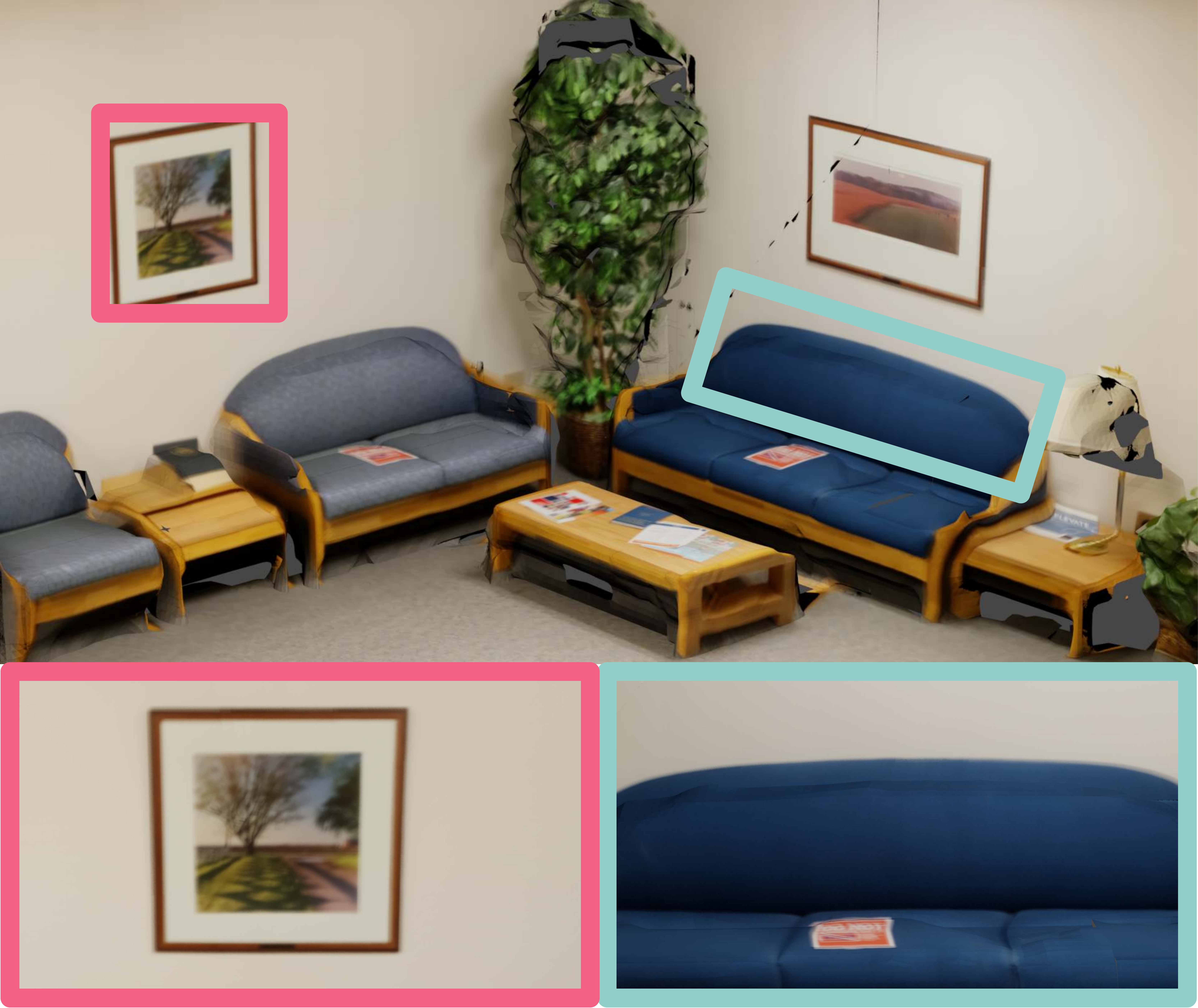}
        \captionsetup{width=\linewidth}
        \caption{L2Avg.}
        \label{fig:init comparison, l2 avg}
    \end{subfigure}%
    \hspace{0.01\columnwidth}
    \begin{subfigure}{0.45\columnwidth}
    \centering
        \includegraphics[width=\linewidth]{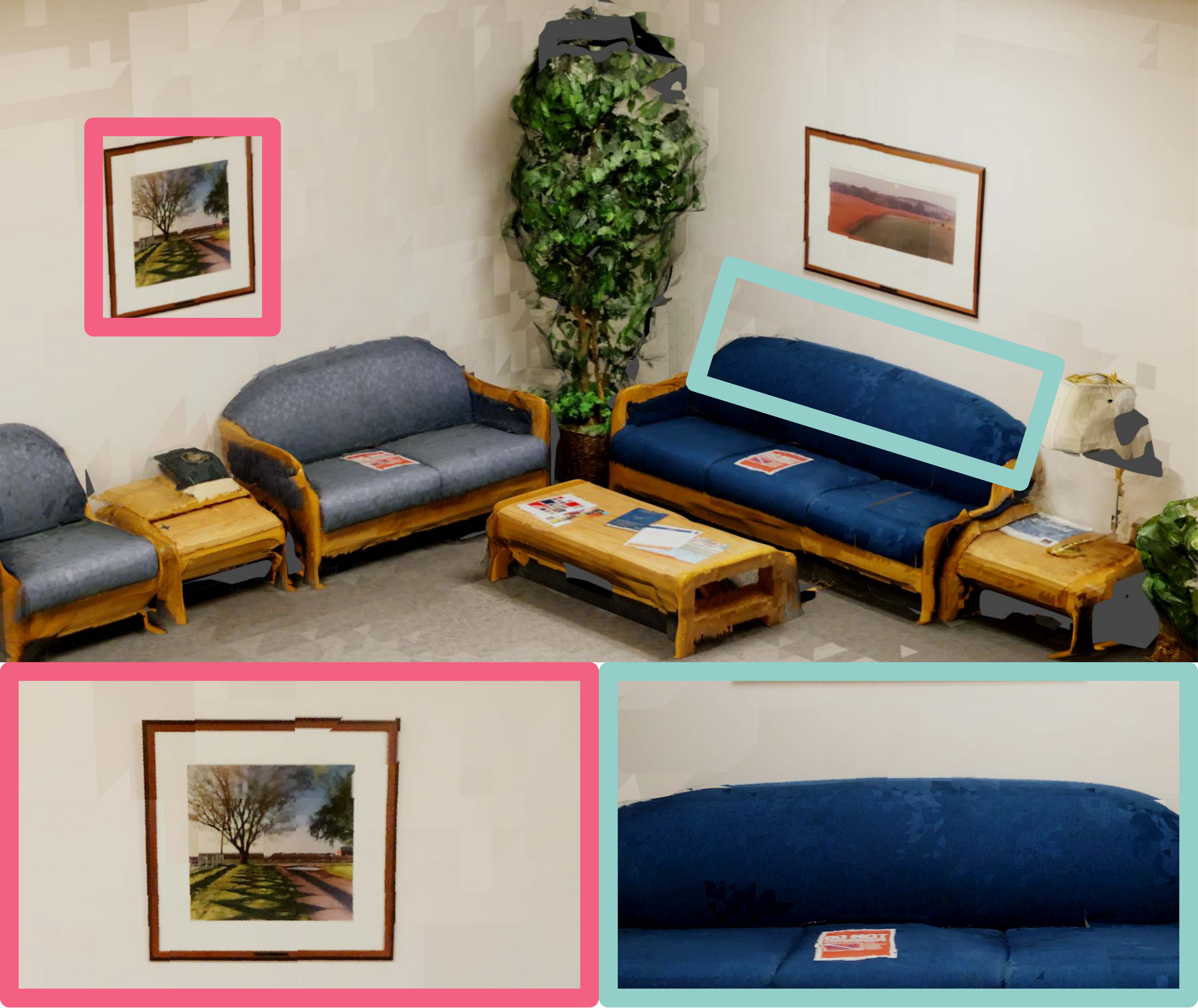}
        \captionsetup{width=\linewidth}
        \caption{$\cT$.}
        \label{fig:init comparison, mrf}
    \end{subfigure}%
    \caption{
    \textbf{Initialization comparison.}
    \textbf{(a)} We use PyTorch3D's rendering pipeline~\cite{ravi2020pytorch3d} to project each pixel of every RGB image back to the texture. The color of each pixel in the texture is the average of all colors that project to it. This texture minimizes the $\mathcal{L}_2$ loss of the difference between the rendered and the ground truth images. We dub it L2Avg.
    \textbf{(b)} $\cT$ from~\secref{sec: atlas gen} permits to maintain details. The seam artifacts will be optimized out using \texttt{TexSmooth} (\secref{sec: tex smooth}).
    Besides  over-smoothness, without taking into account misalignments of geometry and camera poses, L2Avg produces texture that overfits to available views,~\eg,~the sofa's blue colors are painted onto the wall.
    }
    \label{fig: init comparison}
\end{minipage}%
\end{minipage}
\end{figure}

However, despite those guarantees, BFF still requires meshes to have a manifold connectivity.
While we augment work by~\cite{Sawhney2018BoundaryFF} using vertex duplication to circumvent this restriction, flattening may still result in overlapping regions as illustrated in \figref{fig:overlap}. To fix this and uniquely assign a triangle to a position in the texture, we perform overlap detection as discussed next.

\noindent\textbf{Overlap Detection:} Overlap detection operates on flattened and possibly overlapping triangle meshes like the one illustrated in \figref{fig: mesh flatten overlap}. Our goal is to assign triangles to different planes. Upon re-packing the triangles assigned to different planes, we  obtain the non-overlapping triangle mesh illustrated in \figref{fig: mesh flatten non-overlap}. 

In order to not break the triangle mesh at a random position and end up with many individual triangles, \ie, in order to maintain large triangle connectivity, we formulate this problem using a Markov Random Field (MRF). 
Formally, let the discrete variable $y_i\in\cY = \{1, \dots, |\cY|\}$ denote the discrete plane index that the $i$-th triangle $\texttt{Tri}_i$ is assigned to. Hereby, $|\cY|$ denotes the maximum number of planes which is identical to the maximum number of triangles that overlap initially at any one location. We obtain the triangle-plane assignment $y^\ast =  (y_1^\ast, \dots, y_{|M|}^\ast)$ for all $|M|$ triangles by addressing
\begin{equation}
y^\ast = \arg\max_y \sum_{i=1}^{|M|} \phi_i(y_i) + \sum_{(i,j)\in\cA\cup\cO} \phi_{i,j}(y_i,y_j),
\label{eq:MRFPlane}
\end{equation}
where $\cA$ and $\cO$ are sets of triangle index pairs which are adjacent and overlapping respectively. Here, $\phi_i(\cdot)$ denotes triangle $\texttt{Tri}_i$'s priority over $\cY$ when considering only its \textit{local} information, while $\phi_{i,j}(\cdot)$ refers to $\texttt{Tri}_i$ and $\texttt{Tri}_j$'s joint preference on their assignments.
\equref{eq:MRFPlane} is solved with belief propagation~\cite{Globerson2007FixingMC}.

Intuitively, by addressing the program given in \equref{eq:MRFPlane} we want a different plane index for overlapping triangles, while encouraging mesh $M$'s adjacent triangles to be placed on the same plane. 
To achieve this we use
\begin{align}
    \hspace{-0.3cm}\phi_i(y_i) &= 
        \begin{cases}
            1.0,  &\text{if } y_i = \min \cY_{i, \text{non-overlap}} \\
            0.0,  &\text{otherwise} 
        \end{cases}, \text{~and} \label{eq: overlap unary}\\
    \hspace{-0.3cm}\phi_{i,j}(y_i, y_j) &= 
        \begin{cases}
            \mathbbm{1}\{ y_i = y_j \}, &\text{if } (i, j) \in \cA \\
            \mathbbm{1}\{ y_i \neq y_j \}, &\text{if } (i, j) \in \cO
        \end{cases}.
\end{align}
Here, $\mathbbm{1}\{\cdot\}$ denotes the indicator function and
$\cY_{i, \text{non-overlap}}$ %
contains all plane indices where $\texttt{Tri}_i$ has no overlap with others. Intuitively, \equref{eq: overlap unary} encourages to assign the minimum of such indices to $\texttt{Tri}_i$. %

As fast MRF optimizers remove most overlaps but don’t provide guarantees, we add a light post-processing to manually assign the remaining few overlapping triangles to separate planes. This guarantees overlap-free results.
As mentioned before, after having identified the plane assignment $y^\ast$ for each triangle we use a bin packing to uniquely assign each triangle to a position in the texture.
Conversely, for every texture coordinate $u,v$ we obtain a unique triangle index 
\begin{equation}
    i = G(u,v). 
    \label{eq:trianglefromcoord}
\end{equation}
A qualitative result is illustrated in \figref{fig: mesh flatten non-overlap}. 
Next, we identify the image which should be used to texture each triangle.

\noindent\textbf{\myuline{2) Textures from Triangle-Image Assignments:}} 
Our goal is to identify a suitable frame  $I_{t_i}$, $t_i\in\{1, \dots, T+1\}$, for each triangle $\texttt{Tri}_i$, $i\in\{1, \dots, |M|\}$. Note that the $(T+1)$-th option $I_{T+1}$ %
refers to an empty texture.  We compute the texture assignments $\bt^\ast=(t_1^\ast, \dots, t_{|M|}^\ast)$  using  a purely local optimization:
\begin{equation}
    \bt^\ast = \argmax_\bt \sum_{i=1}^{|M|} \psi_i(t_i). %
    \label{eq:MRFTex}
\end{equation}
Here $\psi_i$ captures \textit{unary} cues.
Note, we also studied \textit{pairwise} cues but did not observe significant improvements. Please see the Appendix for more details. Due to better efficiency, we therefore only consider unary cues. %
Intuitively, we want the program given in \equref{eq:MRFTex} to encourage triangle-image assignment to be `best' for each triangle $\texttt{Tri}_i$. 
We describe the unary cues to do so next. %

\begin{figure*}[t]
    \centering
    \includegraphics[width=\textwidth]{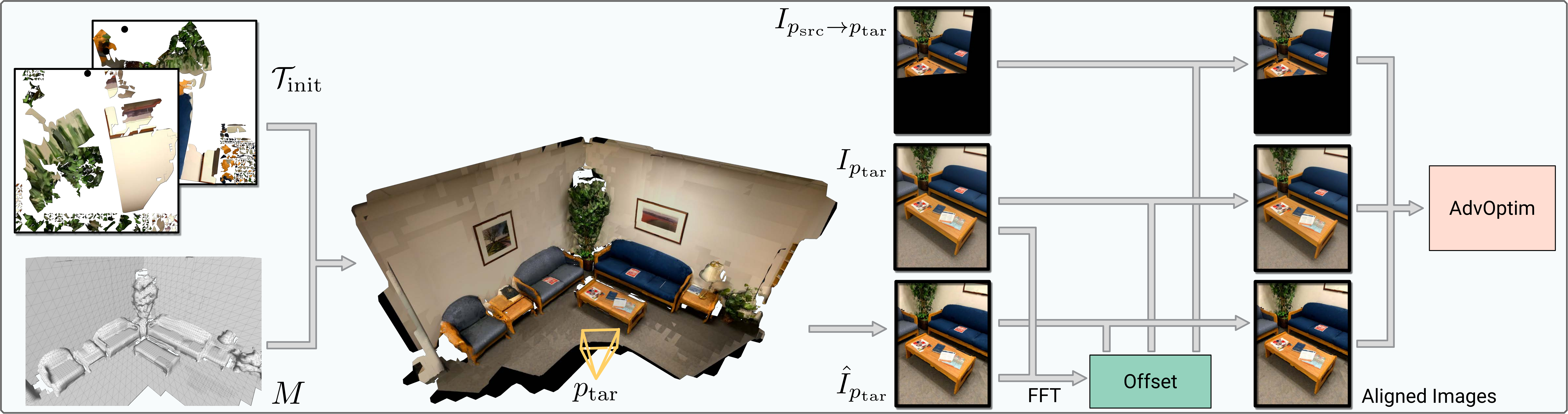}
    \caption{\textbf{Texture smoothing \texttt{TexSmooth}} (\secref{sec: tex smooth}).
    We utilize adversarial optimization (\texttt{AdvOptim})~\cite{Huang2020AdversarialTO} to refine the texture $\cT$ from~\secref{sec: atlas gen}. Differently:
    1) We initialize with $\cT$.
    2) To resolve the issue of misalignment between rendering and ground truth (GT), we  integrate an alignment module based on the fast Fourier transform (FFT). %
    }
    \label{fig: tex adv refine}
\end{figure*}

\noindent\textbf{Unary Potentials $\psi_i(t_i)$} for each pair of triangle $\texttt{Tri}_i$ and frame $I_{t_i}$ are 
\begin{align}
  \psi_i(t_i) &= 
  \begin{cases}
       \psi_i^{\texttt{C}}(t_i), &\text{if } \psi_i^{\texttt{V}}(t_i) = 1 \\
       -\infty, &\text{otherwise}
  \end{cases},
\end{align}
where $\psi_i^{\texttt{V}}(t_i)$ and $\psi_i^{\texttt{C}}(t_i)$ represent validity check and potentials from cues respectively. Concretely, we use
\begin{align}
  \psi_i^{\texttt{V}}(t_i) &= \mathbbm{1}\{ I_{t_i} \in \mathcal{S}_i^\texttt{V} \}, \\
  \psi_i^{\texttt{C}}(t_i) &= \omega_{1} \cdot \psi_i^{\cue{1}}(t_i) + \omega_2 \cdot \psi_i^{\cue{2}}(t_i) + \omega_3 \cdot \psi_i^{\cue{3}}(t_i) \label{eq: tex mrf unary},
\end{align}
where $\mathcal{S}_i^\texttt{V}$ denotes the set of valid frames for $\texttt{Tri}_i$ and $\omega_1, \omega_2, \omega_3$ represent weights for potentials $\psi_i^{\cue{1}}, \psi_i^{\cue{2}}, \psi_i^{\cue{3}}$. We discuss each one next:

\noindent $\bullet$~\textbf{Validity} ($\texttt{V}$). To assess whether frame  $I_{t_i}$ is valid for $\texttt{Tri}_i$, we  check the visibility of $\texttt{Tri}_i$ in $I_{t_i}$. We approximate this by checking visibility of $\texttt{Tri}_i$'s three vertices as well as its centroid. %
Concretely, we transform the vertices and centroid from world coordinates to the normalized device coordinates of the $t_i$-th camera. %
If all vertices and centroid are visible, \ie, their coordinates are in the interval $[-1,1]$, we add frame $I_{t_i}$ to the set $\mathcal{S}_i^\texttt{V}$ of valid frames for triangle $\texttt{Tri}_i$.

\noindent $\bullet$~\textbf{Triangle area} ($\cue{1}$). Based on a camera's  pose $p_{t_i}$, a triangle's area changes. %
The larger the area, the more detailed is the information for $\texttt{Tri}_i$  in frame $I_{t_i}$. We encourage to assign $\texttt{Tri}_i$ to frames $I_{t_i}$ with large area by defining $\psi_i^{\cue{1}}(t_i) = \texttt{Area}_{t_i}(\texttt{Tri}_i)$ and set $\omega_1 > 0$.

\noindent $\bullet$~\textbf{Discrepancy between $z$-buffer and actual depth} ($\cue{2}$).
For a valid frame $I_{t_i} \in \mathcal{S}_i^\texttt{V}$, a triangle's vertices and its centroid project to valid image coordinates.
We compute the discrepancy between: 
1) the depth  from frame $I_{t_i}$ at the image coordinates of the vertices and centroid;
2) the depth of vertices and centroid in the camera's coordinate system. 
We set $\psi_i^{\cue{2}}(t_i)$ to be the sum of absolute value differences between both depth estimates while using $\omega_2 < 0$.

\noindent $\bullet$~\textbf{Perceptual consistency} ($\cue{3}$). Due to diverse illumination, triangle $\texttt{Tri}_i$'s appearance changes across frames. %
Intuitively, we don't want to assign a texture to $\texttt{Tri}_i$ using a frame that contains colors that deviate drastically from other frames. 
Concretely, we first average all triangle's three vertices color values across all valid frames, \ie, across all $ I_{t_i} \in \mathcal{S}_i^\texttt{V}$.
We then compare this global average to the local average obtained independently for the three vertices of every valid frame $I_{t_i} \in \mathcal{S}_i^\texttt{V}$ using an absolute value difference. 
We require $\omega_3 < 0$.

\noindent\textbf{\myuline{3) Color Transfer:}} 
Given the inferred triangle-frame assignments $\bt^\ast$ we  complete the texture $\cT$ by transferring RGB data from image $I_{t_i^\ast}$ for $\texttt{Tri}_i, i\in\{1, \dots, |M|\}$. For this we leverage the camera pose $p_{t_i^\ast}$ which permits to transform the texture coordinates $(u,v)$ of locations within $\texttt{Tri}_i$ to corresponding image coordinates $(a, b)$ in texture $I_{t_i^\ast}$ via the mapping $F : \mathbb{R}^2 \rightarrow \mathbb{R}^2$, \ie, 
\begin{equation}
    (a,b) = F(u,v,t_i^\ast,p_{t_i^\ast}).
    \label{eq:Mapping}
\end{equation}
Intuitively, given the $(u,v)$ coordinates on the texture in a coordinate system which is local to the triangle $\texttt{Tri}_i$, and given the camera pose $p_{t_i^\ast}$ used to record image $I_{t_i^\ast}$, the mapping $F$ retrieves the  image coordinates $(a,b)$ corresponding to texture coordinate $(u,v)$. Using this mapping, we obtain the texture $\cT$ at location $(u,v)$, \ie, $\cT(u,v)$, from the image data $I_{t_i^\ast}(a,b)\in\mathbb{R}^3$ via
\begin{equation}
    \cT(u,v) = I_{t_i^\ast}(F(u,v,t_i^\ast, p_{t_i^\ast})).
\end{equation}

\noindent Note, because of the overlap detection, we obtain a unique triangle index $i=G(u,v)$ for every $(u,v)$ coordinate from \equref{eq:trianglefromcoord}.
Having transferred RGB data for all coordinates within all triangles results in the texture $\cT\in\mathbb{R}^{H\times W\times 3}$, which we compare to standard L2 averaging initialization in \figref{fig: init comparison}. We next refine this texture via adversarial optimization. We observe that this initialization $\cT$ is crucial to obtain  high-quality textures, which we will show in~\secref{sec:exp}.

\csubsection{Texture Smoothing (\texttt{TexSmooth})}\label{sec: tex smooth}

As can be seen in~\figref{fig:init comparison, mrf}, the texture $\cT$  contains seams that affect visual quality. To reconstruct a  seamless texture $\cTfinal$, we extend  recent adversarial optimization (\texttt{AdvOptim}). %
Different from prior work~\cite{Huang2020AdversarialTO}  which initializes with blank (paper) or averaged (code release\footnote{\label{footnote:advtex equ}\url{https://github.com/hjwdzh/AdversarialTexture}}) textures, we initialize with $\cT$.
Also, we find \texttt{AdvOptim} doesn't handle common camera pose and geometry misalignment well. To resolve this, we  develop an efficient  alignment module. This is depicted in~\figref{fig: tex adv refine} and will be detailed next.

\noindent\textbf{\myuline{Smoothing with Adversarial Optimization:}} 
To optimize the texture, \texttt{AdvOptim} iterates over camera poses. 
When optimizing for a specific target camera pose $p_\text{tar}$, \texttt{AdvOptim} uses three images: 1) the ground truth image $I_{p_\text{tar}}$ of the target camera pose $p_\text{tar}$; 2) a rendering  $\hat{I}_{p_\text{tar}}$ for the target camera pose $p_\text{tar}$ from the texture map $\cTfinal$;  and 3) a re-projection from another camera pose $p_\text{src}$'s ground truth image, which we refer to as  $I_{p_\text{src}\rightarrow p_\text{tar}}$. It then optimizes by minimizing an $\mathcal{L}_1$  plus a conditional adversarial loss.
However, we find \texttt{AdvOptim} to struggle with alignment errors due to %
inaccurate geometry.
Therefore, we integrate an efficient  alignment operation into \texttt{AdvOptim}.
Instead of directly using the input images, we first compute a 2D offset $(\Delta h_{p_\text{tar}}, \Delta w_{p_\text{tar}})$ between $I_{p_\text{tar}}$ and $\hat{I}_{p_\text{tar}}$, which we apply to align $I_{p_\text{tar}}$ and $\hat{I}_{p_\text{tar}}$ as well as $I_{p_\text{src}\rightarrow p_\text{tar}}$ via
\begin{align}
    I^\texttt{A} \doteq \texttt{Align} (I, (\Delta h, \Delta w)),
\end{align}
where $I^\texttt{A}$ marks aligned images.
We then use the three aligned images as input:

\begin{align}\label{eq: adv optim loss}
    \mathcal{L} =  &\lambda \Vert I_{p_\text{tar}}^\texttt{A} - \hat{I}_{p_\text{tar}}^\texttt{A} \Vert_1 
    + \mathbb{E}_{I_{p_\text{tar}}^\texttt{A}, I_{p_\text{src}\rightarrow p_\text{tar}}^\texttt{A}} \left[ \log D(I_{p_\text{src}\rightarrow p_\text{tar}}^\texttt{A} \vert I_{p_\text{tar}}^\texttt{A} ) \right] \nonumber \\
    &
    + \mathbb{E}_{I_{p_\text{tar}}^\texttt{A}, \hat{I}_{p_\text{tar}}^\texttt{A}} \left[ \log ( 1 - D (\hat{I}_{p_\text{tar}}^\texttt{A} \vert I_{p_\text{tar}}^\texttt{A}) )  \right].
\end{align}
Here, $D$ is a convolutional deep-net based discriminator.
When using the unaligned image $I$ instead of $I^\texttt{A}$,~\equref{eq: adv optim loss} reduces to the vanilla version in~\cite{Huang2020AdversarialTO}.
We now discuss a fast way to align images.

\begin{figure}[!t]
\begin{minipage}[t]{\columnwidth}
\centering
\begin{minipage}{.62\columnwidth}
    \centering
    \includegraphics[width=\columnwidth]{./fig/fft/two_row_fft_offset.pdf}
    \caption{
      \textbf{Alignment with fast Fourier transformation (FFT)} (\secref{sec: tex smooth}). We show results for $\cT$ (\secref{sec: atlas gen}) and L2Avg (\figref{fig: init comparison}) in top and bottom rows.
      \textbf{(a):} ground-truth (GT);
      \textbf{(b) and (e):} texture rendering with (a)'s corresponding camera;
      \textbf{(c) and (d):} difference between (a) and (b);
      \textbf{(f) and (g):} difference between (a) and (e).
      The top row: FFT successfully aligns GT image and rendering from $\cT$.
      Within expectation, there is almost no misalignment for the texture L2Avg as it overfits to available views (\figref{fig: init comparison}).}
    \label{fig:fft offset}
\end{minipage}
\hspace{0.01\columnwidth}
\begin{minipage}{.35\columnwidth}
    \centering
    \captionsetup[subfigure]{aboveskip=1pt}
    \begin{subfigure}{0.3\columnwidth}
        \centering
        \includegraphics[width=\columnwidth]{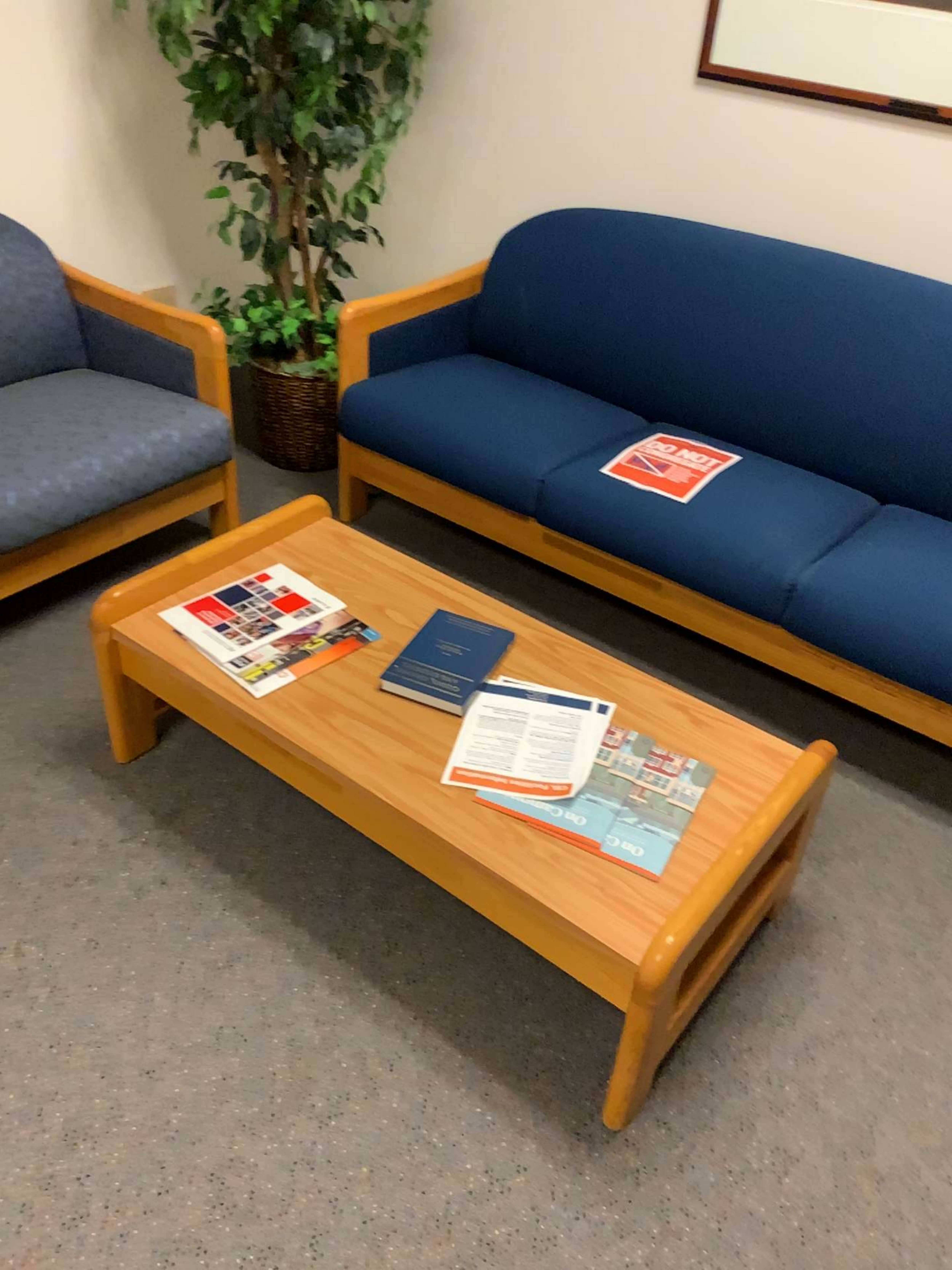}
        \captionsetup{width=\columnwidth}
        \caption{}
        \label{fig: eval align gt}
    \end{subfigure}
    \hspace{0.01\columnwidth}
    \begin{subfigure}{0.3\columnwidth}
        \centering
        \includegraphics[width=\columnwidth]{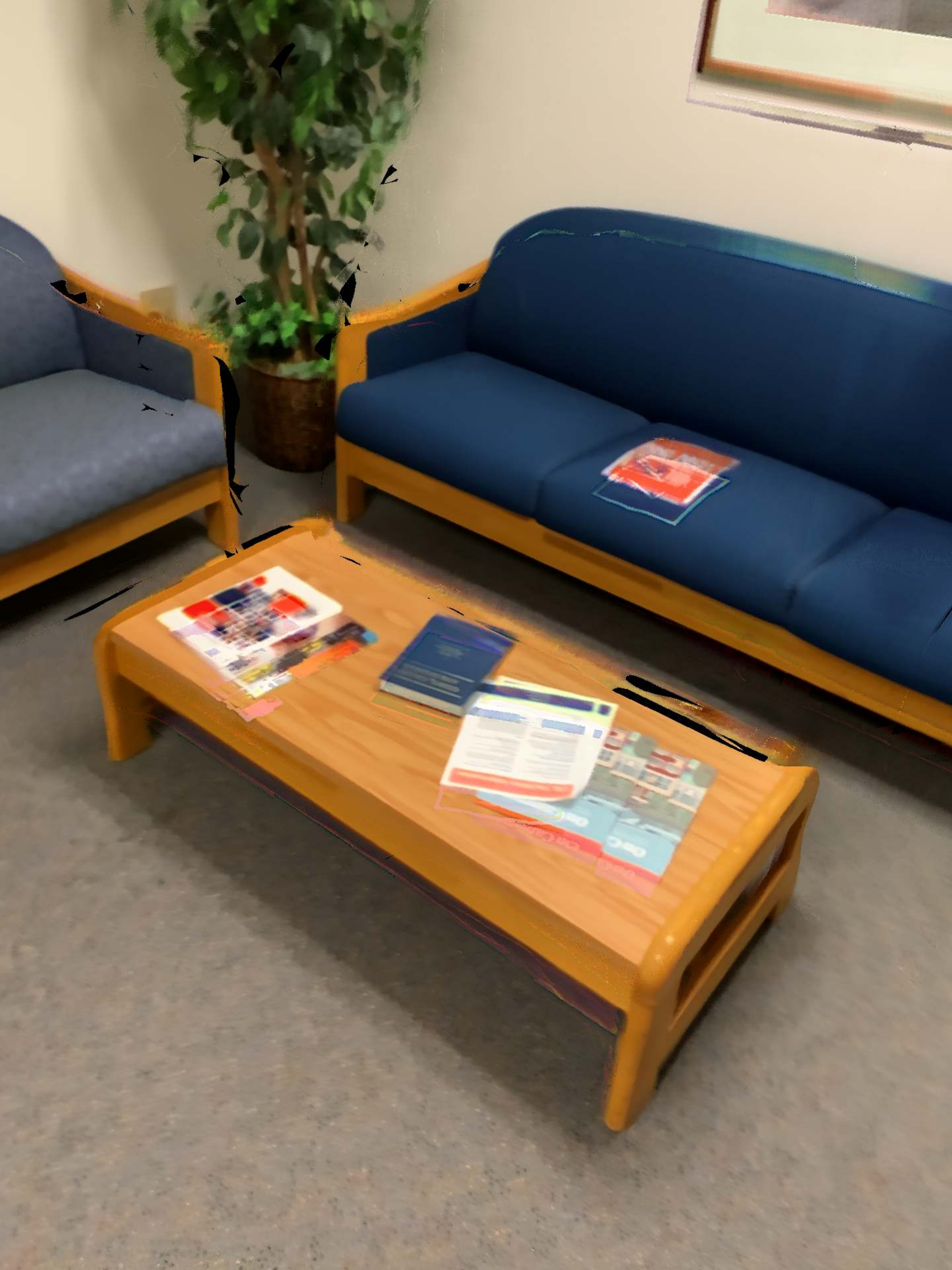}
        \captionsetup{width=\columnwidth}
        \caption{}
        \label{fig: eval align adv}
    \end{subfigure}%
    \hspace{0.01\columnwidth}
    \begin{subfigure}{0.3\columnwidth}
        \centering
        \includegraphics[width=\columnwidth]{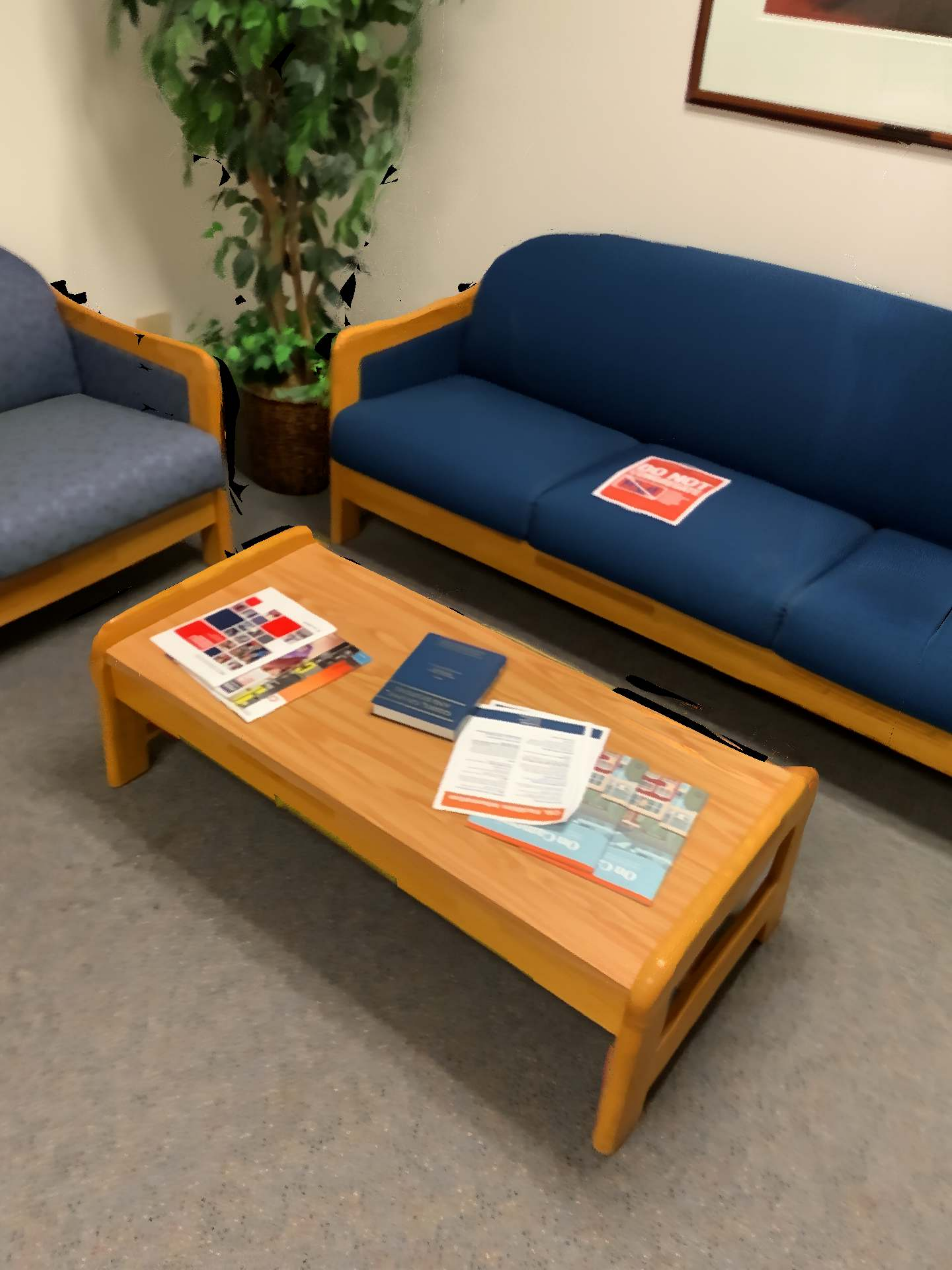}
        \captionsetup{width=\columnwidth}
        \caption{}
        \label{fig: eval align adv offset}
    \end{subfigure}%
    \hfill
    \caption{
    \textbf{Alignment is important for evaluation (\secref{sec: exp setup}).}
    Clearly, (c) is more desirable than (b).
    However, before alignment, LPIPS yields 0.3347 and 0.4971 for (a)-(b) and (a)-(c) pairs respectively. This is misleading as lower LPIPS indicates higher quality.
    After alignment, LPIPS produces 0.3343 and 0.2428 for the same pairs, which provides correct signals for evaluation.
    }
    \label{fig: eval align}
\end{minipage}%
\end{minipage}
\end{figure}

\noindent\textbf{\myuline{Alignment with Fourier Transformation:}} 
To align ground truth $I_{p_\text{tar}}$ and rendering $\hat I_{p_\text{tar}}$, one could use na\"ive grid-search to find the offset which results in the minimum difference of the shifted images. %
However, such a grid-search is prohibitively costly  during an iterative optimization, especially with high-resolution images (\eg,~we use a resolution up to 1920$\times$1440).
Instead, %
we use the fast Fourier transformation (FFT)  to complete the job~\cite{AnutaGeo1970}.
Specifically, 
given a misaligned image pair of $I \in \mathbb{R}^{h\times w\times 3}$ and $\hat{I} \in \mathbb{R}^{h\times w \times 3}$, %
we compute for every channel the maximum correlation via 
\begin{align}
    \argmax\limits_{(i, j)}\; \texttt{FFT}^{-1} \left( \texttt{FFT}(I) \cdot \overline{\texttt{FFT}(\hat{I})} \,\right)[i, j]. 
\end{align}
Here, $\texttt{FFT}(\cdot)$ represents the fast Fourier transformation  %
while $\texttt{FFT}^{-1}(\cdot)$ denotes its inverse and  $\overline{\texttt{FFT}(\hat{I})}$ refers to the complex conjugate.
After decoding the maximum correlation response and averaging across channels, we obtain the 
final offset $(\Delta h, \Delta w)$.
As can be seen in~\figref{fig:fft offset}(d), the offset $(\Delta h, \Delta w)$ is very accurate.
Moreover, the  computation finishes in around 0.4 seconds even for $1920\times1440$-resolution images. %
Note, we don't need to maintain gradients for  $(\Delta h, \Delta w)$, since the offset is only used to shift images and not to backpropagate through it.

\csection{Experiments}
\label{sec:exp}

\csubsection{Experimental Setup}\label{sec: exp setup}

\noindent \textbf{Data Acquisition.} We use a 2020 iPad Pro and develop an iOS app to acquire the RGBD images $I_t$, camera pose $p_{t}$, and scene mesh $M$ via Apple's ARKit~\cite{arkit2021}. %

\noindent \textbf{UofI Texture Scenes.} We collect a dataset of 11 scenes: four indoor and seven outdoor scenes. This dataset consists of a total of 2807 frames, of which 91, 2052, and 664 are of resolution $480\times 360$, $960\times720$, and $1920\times 1440$ respectively. 
For each scene, we use 90\% of its views for optimization and the remainder for evaluation. 
In total, we have 2521 training frames and 286 test frames.
This setting is more challenging than prior work where~\cite{Huang2020AdversarialTO} ``select(s) 10 views uniformly distributed from the scanning video'' for evaluation while using up to thousands of frames  for texture generation. %
On average, the angular differences between test set view directions and their nearest neighbour in the training sets are 2.05$^\circ$ (min 0.85$^\circ$/max 13.8$^\circ$). Angular distances are computed following~\cite{Huynh2009MetricsF3}. Please see Appendix
for scene-level statistics.

\noindent \textbf{Implementation.} We compare to five baselines for texture generation: L2Avg, ColorMap~\cite{Zhou2014ColorMO}, TexMap~\cite{Fu2018TextureMF}, MVSTex~\cite{Waechter2014LetTB}, and AdvTex~\cite{Huang2020AdversarialTO}.
For ColorMap, TexMap, and MVSTex, we use their official implementations.\footnote{\scriptsize ColorMap: \url{https://github.com/intel-isl/Open3D/pull/339}; TexMap: \url{https://github.com/fdp0525/G2LTex}; MVSTex: \url{https://github.com/nmoehrle/mvs-texturing}}
For  \texttt{AdvOptim} (\secref{sec: tex smooth}) used in both AdvTex and ours, we re-implement a PyTorch~\cite{Paszke2019PyTorchAI} version based on their official release in TensorFlow~\cite{Abadi2016TensorFlowAS}.\cref{footnote:advtex equ}
We evaluate AdvTex with two different initializations: 1) blank textures as stated in the paper (AdvTex-B); 2) the initialization used in the official code release (AdvTex-C).
We run \texttt{AdvOptim} using the Adam optimizer~\cite{Kingma2015AdamAM}.
See Appendix
for more details.
For our \texttt{TexInit} (\secref{sec: atlas gen}), we use a generic set of weights across all scenes: $\omega_1 = 1e^{-3}$ (triangle area), $\omega_2 = -10$ (depth discrepancy), and $\omega_3 = -1$ (perception consistency), which makes cue magnitudes roughly similar.

On a 3.10GHz Intel Xeon Gold 6254 CPU, ColorMap takes less than two minutes to complete  while TexMap's running time ranges from 40 minutes to 4 hours.
MVSTex can be completed in no more than 10 minutes.
Our $\cT$ (\secref{sec: atlas gen}) completes in two minutes. Additionally, the \texttt{AdvOptim} takes around 20 minutes for 4000 iterations to complete with an Nvidia RTX A6000 GPU.

\noindent \textbf{Evaluation metrics.}
To assess the efficacy of the method, we study the quality of the texture from two perspectives: perceptual quality and sharpness.
1) For perceptual quality, we assess the
 similarity between rendered and collected ground-truth views using the Structural Similarity Index Measure (SSIM)~\cite{Wang2004ImageQA} and the Learned Perceptual Image Patch Similarity (LPIPS)~\cite{Zhang2018TheUE}.
2) For sharpness, we consider measurement $S_3$~\cite{Vu2012bfSA} and the norm of image gradient (Grad) following~\cite{Huang2020AdversarialTO}.
Specifically, for each pixel, we compute its $S_3$ value, whose difference between the rendered and ground truth (GT) is used for averaging across the whole image. A similar procedure is applied to Grad.
For all four metrics, we report the mean and standard deviation  across 11 scenes.

\noindent \textbf{Alignment in evaluation.} As can be seen in~\figref{fig: eval align}, evaluation will be misleading if we do not align images during evaluation. Therefore, we propose the following procedure:
1) for each method, we align the rendered image and the GT using an \texttt{FFT} (\secref{sec: tex smooth});
2) to avoid various resolutions caused by different methods, we crop out the maximum common area across methods.
3) we then compute metrics on those cropped regions.
The resulting comparison is fair as all methods are evaluated on the same number of pixels and aligned content.

\csubsection{Experimental Evaluation}\label{sec: exp eval}

\noindent \textbf{Quantitative evaluation.} \tabref{tab: qunatitative} reports aggregated results on all 11 scenes.
The quality of our texture $\cTfinal$ (Row 6) outperforms baselines on LPIPS, $S_3$ and Grad, confirming the effectiveness of the proposed pipeline. 
Specifically, we improve LPIPS by 7.8\% from 0.335 ($2^\text{nd}$-best) to 0.309, indicating high perceptual similarity.
Moreover, $\cTfinal$ maintains sharpness as we improve $S_3$ by 11.1\% from 0.135 ($2^\text{nd}$-best) to 0.120 and Grad from 7.171 ($2^\text{nd}$-best) to 6.871. 
Regarding SSIM, we find it to favor L2Avg in almost all scenes
(see Appendix)
which aligns with the findings in~\cite{Zhang2018TheUE}.\\ %

\noindent \textbf{Ablation study.} We verify the design choices of \texttt{TexInit} and \texttt{TexSmooth} in~\tabref{tab: ablation}.
\textbf{1) \texttt{TexSmooth} is required:} we directly evaluate $\cT$ and $1^\text{st}$~\vs~$4^\text{th}$ row confirms the performance drop: -0.092 (SSIM), +0.033 (LPIPS), +0.021 ($S_3$), and +1.221 (Grad).
\textbf{2) $\cT$ is needed:} we replace $\cT$ with L2Avg as it performs better than ColorMap and TexMap in~\tabref{tab: qunatitative} and still incorporate \texttt{FFT} into \texttt{AdvOptim}.
We observe  inferior performance: -0.010 (SSIM), +0.023 (LPIPS), +0.010 ($S_3$) in $2^\text{nd}$~\vs~$4^\text{th}$ row.
\textbf{3) Alignment is important:} we use the vanilla \texttt{AdvOptim} but initialize with $\cT$. As shown in~\tabref{tab: ablation}'s $3^\text{rd}$~\vs~$4^\text{th}$ row, the texture quality drops by -0.043 (SSIM), +0.037 (LPIPS), +0.005 ($S_3$), and +0.373 (Grad).\\

\begin{table}[t]
\renewcommand{\arraystretch}{0.9}
\begin{adjustwidth}{0.0cm}{}
\captionsetup{width=\linewidth}
\caption{
\textbf{Aggregated quantitative evaluation on UofI Texture Scenes}. We report results in the form of $\texttt{mean}{\scriptsize\pm\texttt{std}}$. 
Please see~\figref{fig: qualitative} for qualitative texture comparisons and  Appendix~\ref{supp sec: scene-level quant} for scene-level quantitative results.
}
\label{tab: qunatitative}
\begin{adjustbox}{width=0.7\columnwidth,center}
\renewcommand\theadfont{}
\centering
\setlength{\tabcolsep}{3pt}
{
\small
\begin{tabular}{lcrrrr} 
\toprule
 &  & SSIM$\uparrow$  & LPIPS$\downarrow$ & $S_3\downarrow$ & Grad$\downarrow$ \\
\midrule
1 & L2Avg & \textbf{0.610}{\scriptsize$\pm$0.191}  & 0.386{\scriptsize$\pm$0.116}  & 0.173{\scriptsize$\pm$0.105}  & 7.066{\scriptsize$\pm$4.575} \\
2 & ColorMap  & 0.553{\scriptsize$\pm$0.193}  & 0.581{\scriptsize$\pm$0.132}  & 0.234{\scriptsize$\pm$0.140}  & 7.969{\scriptsize$\pm$5.114} \\
3 & TexMap & 0.376{\scriptsize$\pm$0.113}  & 0.488{\scriptsize$\pm$0.097}  & 0.179{\scriptsize$\pm$0.062}  & 8.918{\scriptsize$\pm$4.174}   \\
\midrule

4 & MVSTex & 0.476{\scriptsize$\pm$0.164}  & 0.335{\scriptsize$\pm$0.086}  & 0.139{\scriptsize$\pm$0.047}  & 8.198{\scriptsize$\pm$3.936}  \\

5-1 & AdvTex-B & 0.495{\scriptsize$\pm$0.174}  & 0.369{\scriptsize$\pm$0.092}  & 0.148{\scriptsize$\pm$0.047}  & 8.229{\scriptsize$\pm$4.586}  \\
5-2 & AdvTex-C & 0.563{\scriptsize$\pm$0.191}  & 0.365{\scriptsize$\pm$0.096}  & 0.135{\scriptsize$\pm$0.067}  & 7.171{\scriptsize$\pm$4.272}   \\
\midrule
\belowrulesepcolor{highlightRowColor} 
\rowcolor{highlightRowColor}
6 & Ours & 0.602{\scriptsize$\pm$0.189}  & \textbf{0.309}{\scriptsize$\pm$0.086}  & \textbf{0.120}{\scriptsize$\pm$0.058}  & \textbf{6.871}{\scriptsize$\pm$4.342}  \\
\toprule
\end{tabular}
}
\end{adjustbox}
\end{adjustwidth}
\end{table}

\begin{table}[t]
\renewcommand{\arraystretch}{0.9}
\begin{adjustwidth}{0.0cm}{}
\captionsetup{width=\linewidth}
\caption{\textbf{Ablation study}. We report results in the form of $\texttt{mean}{\scriptsize\pm\texttt{std}}$.}
\label{tab: ablation}
\begin{adjustbox}{width=0.7\columnwidth,center}
\renewcommand\theadfont{}
\centering
\setlength{\tabcolsep}{3pt}
{
\small
\begin{tabular}{lcccrrrr} 
\toprule
 & {\scriptsize \makecell{Adv\\Optim}} & {\scriptsize \makecell{FFT\\Align}} & \makecell{$\cT$} & SSIM$\uparrow$  & LPIPS$\downarrow$ & $S_3\downarrow$ & Grad$\downarrow$ \\
\midrule
1 &        &        & \cmark & 0.510{\scriptsize$\pm$0.175}  & 0.342{\scriptsize$\pm$0.060}  & 0.141{\scriptsize$\pm$0.052}  & 8.092{\scriptsize$\pm$4.488}  \\
2 & \cmark & \cmark &        & 0.592{\scriptsize$\pm$0.192}  & 0.332{\scriptsize$\pm$0.102}  & 0.130{\scriptsize$\pm$0.066}  & \textbf{6.864}{\scriptsize$\pm$4.211} \\
3 & \cmark &        & \cmark & 0.559{\scriptsize$\pm$0.196}  & 0.346{\scriptsize$\pm$0.082}  & 0.125{\scriptsize$\pm$0.057}  & 7.244{\scriptsize$\pm$4.359}    \\
\midrule
\belowrulesepcolor{highlightRowColor} 
\rowcolor{highlightRowColor}
4 & \cmark & \cmark & \cmark & \textbf{0.602}{\scriptsize$\pm$0.189}  & \textbf{0.309}{\scriptsize$\pm$0.086}  & \textbf{0.120}{\scriptsize$\pm$0.058}  & 6.871{\scriptsize$\pm$4.342} \\
\toprule
\end{tabular}
}
\end{adjustbox}
\end{adjustwidth}
\end{table}

\begin{figure*}[p]
\centering
    \captionsetup[subfigure]{aboveskip=1pt}
    \begin{subfigure}{\textwidth}
        \centering
        \includegraphics[width=\textwidth]{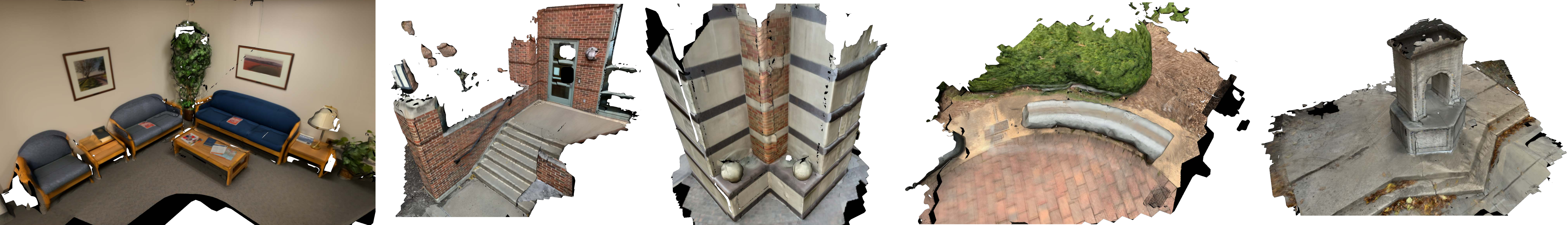}
        \captionsetup{width=\linewidth}
        \caption{\scriptsize \textbf{L2Avg.}}
        \label{fig: qualitative L2Avg}
    \end{subfigure}%
    \hfill
    \begin{subfigure}{\textwidth}
        \centering
        \includegraphics[width=\textwidth]{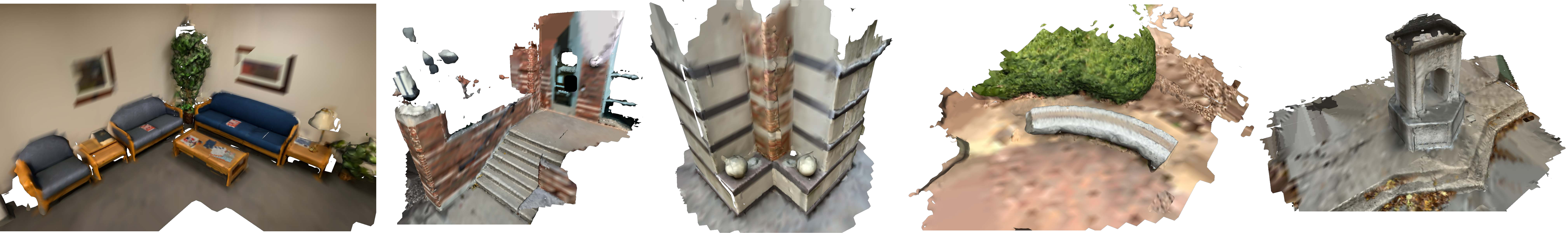}
        \captionsetup{width=\linewidth}
        \caption{\scriptsize \scriptsize \textbf{ColorMap}~\cite{Zhou2014ColorMO}.}
        \label{fig: qualitative ColorMap}
    \end{subfigure}%
    \hfill
    \begin{subfigure}{\textwidth}
        \centering
        \includegraphics[width=\textwidth]{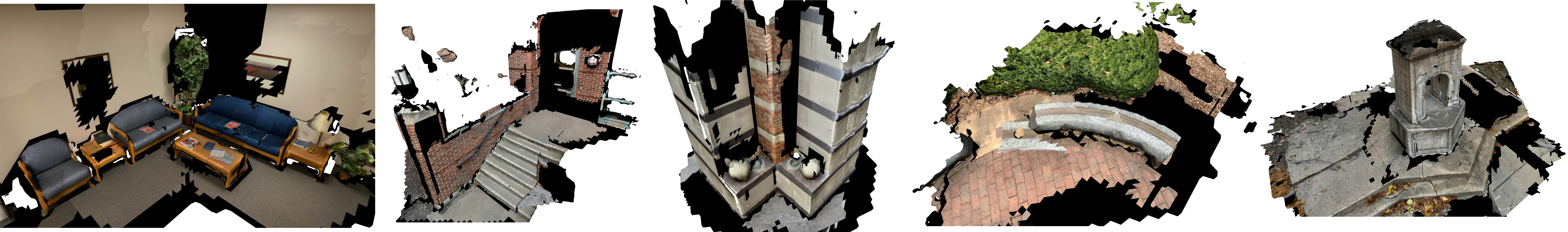}
        \captionsetup{width=\linewidth}
        \caption{\scriptsize \textbf{TexMap}~\cite{Fu2018TextureMF}.}
        \label{fig: qualitative TexMap}
    \end{subfigure}%
    \hfill
    \begin{subfigure}{\textwidth}
        \centering
        \includegraphics[width=\textwidth]{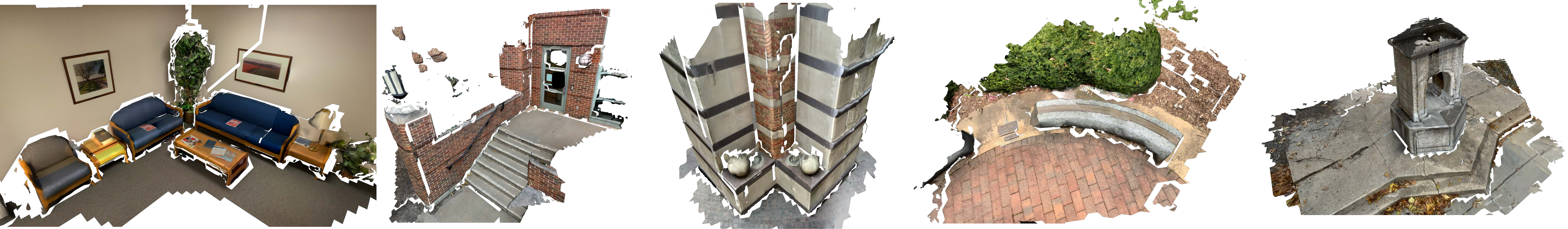}
        \captionsetup{width=\linewidth}
        \caption{\scriptsize \textbf{MVSTex}~\cite{Waechter2014LetTB}. It removes geometry, which is not desirable.}
        \label{fig: qualitative mvstex}
    \end{subfigure}%
    \hfill
    \begin{subfigure}{\textwidth}
        \centering
        \includegraphics[width=\textwidth]{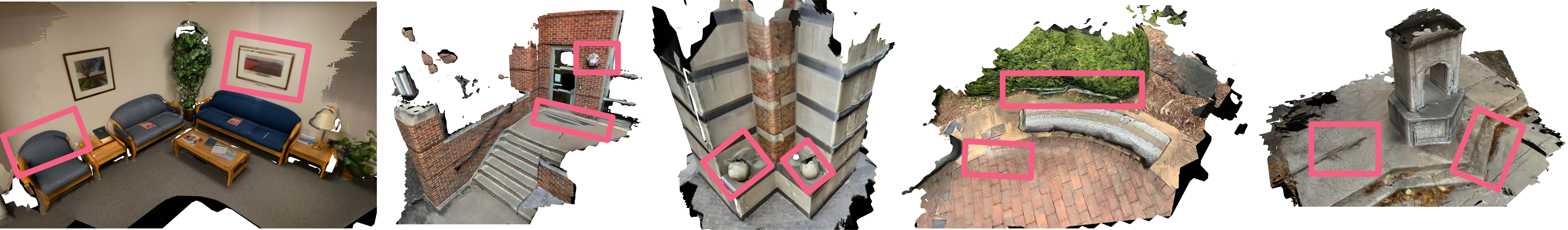}
        \captionsetup{width=\linewidth}
        \caption{\scriptsize \textbf{AdvTex-C}~\cite{Huang2020AdversarialTO}. We highlight artifacts with boxes.
        1) Scene 1: sofa's texture is mapped to the wall and the figure on the wall is broken;
        2) Scene 2: the door's color is mapped to the floor and the brick wall's pattern is mapped to light;
        3) Scene 3: ball's color is projected to brick walls;
        4) Scene 4: bench's color is added to the bush and ground;
        5) Scene 5: the crack breaks and stair's color is on the ground.
        }
        \label{fig: qualitative AdvTex}
    \end{subfigure}%
    \hfill
    \begin{subfigure}{\textwidth}
        \centering
        \includegraphics[width=\textwidth]{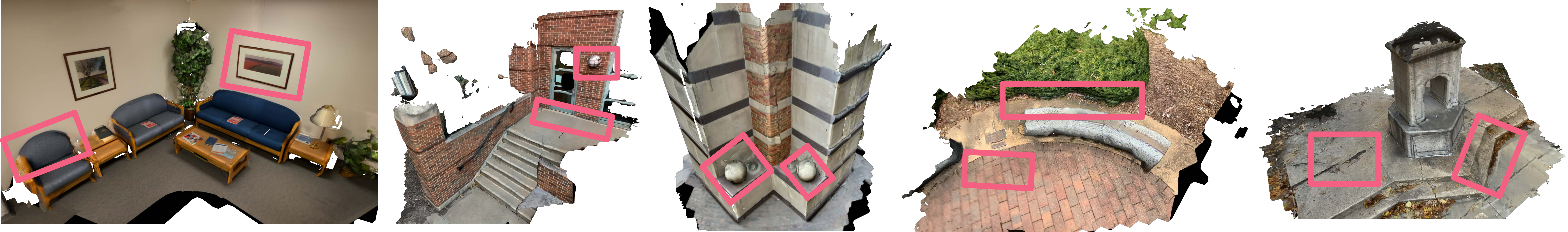}
        \captionsetup{width=\linewidth}
        \caption{\scriptsize \textbf{Ours}. Compared to~\figref{fig: qualitative AdvTex}, our method reduces artifacts.}
        \label{fig: qualitative ours}
    \end{subfigure}%
    \hfill
    \begin{subfigure}{\textwidth}
        \centering
        \includegraphics[width=\textwidth]{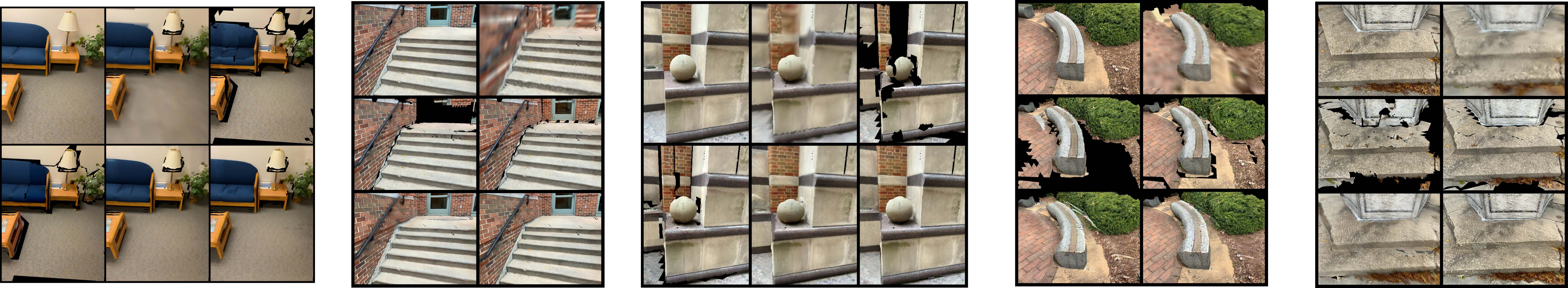}
        \captionsetup{width=\linewidth}
        \caption{\scriptsize
        \textbf{Highlights}. From left to right and top to bottom, we show ground-truth image and renderings at the same camera pose with texture from ColorMap, TexMap, MVSTex, AdvTex-C, and ours respectively.
        Compared to AdvTex-C:
        1) Scene 1: ours produces much more complete and sharper pattern for the sofa;
        2) Scene 2: ours generates sharper cracks on brick walls;
        3) Scene 3: our balls are more complete while AdvTex-C maps the top of the ball to the left brick wall;
        4) Scene 4: AdvTex-C maps the bench to the bush while ours is much cleaner;
        5) Scene 5: the pattern on the ground from ours is much sharper.
        }
        \label{fig: qualitative highlights}
    \end{subfigure}%
    \caption{
      \textbf{Qualitative results on UofI Texture Scenes.}
      For each method, we show results for Scene 1 to 5 from left to right.
      Best viewed in color and zoomed-in.
    }
    \label{fig: qualitative}
\end{figure*}

\begin{table}[t]
\renewcommand{\arraystretch}{0.9}
\begin{adjustwidth}{0.0cm}{}
\captionsetup{width=\linewidth}
\caption{
\textbf{ScanNet results.} We report results in the form of $\texttt{mean}{\scriptsize\pm\texttt{std}}$.
Note, this can't be directly compared to [23]'s Tab.~2: %
while we reserve 10\% views for evaluation, [23] reserves only 10/2011 ($\approx$ 0.5\%), where 2011 is the number of average views per scene.
}
\label{tab: qunatitative scannet}
\begin{adjustbox}{width=0.7\columnwidth,center}
\renewcommand\theadfont{}
\centering
\setlength{\tabcolsep}{3pt}
{
\small
\begin{tabular}{lcrrrr} 
\toprule
 &  & SSIM$\uparrow$  & LPIPS$\downarrow$ & $S_3\downarrow$ & Grad$\downarrow$ \\
\midrule
1-1 & AdvTex-B & 0.534{\scriptsize$\pm$0.074}  & 0.557{\scriptsize$\pm$0.071}  & 0.143{\scriptsize$\pm$0.028}  & 3.753{\scriptsize$\pm$0.730} \\
1-2 & AdvTex-C & 0.531{\scriptsize$\pm$0.074}  & 0.558{\scriptsize$\pm$0.075}  & 0.161{\scriptsize$\pm$0.044}  & 4.565{\scriptsize$\pm$1.399}  \\
\midrule
\belowrulesepcolor{highlightRowColor} 
\rowcolor{highlightRowColor}
2 & Ours  & \textbf{0.571}{\scriptsize$\pm$0.069}  & \textbf{0.503}{\scriptsize$\pm$0.090}  & \textbf{0.127}{\scriptsize$\pm$0.031}  & \textbf{3.324}{\scriptsize$\pm$0.826}    \\
\toprule
\end{tabular}
}
\end{adjustbox}
\end{adjustwidth}
\end{table}

\begin{figure}[t]
\begin{minipage}[t]{\columnwidth}
    \centering
    \includegraphics[width=\textwidth]{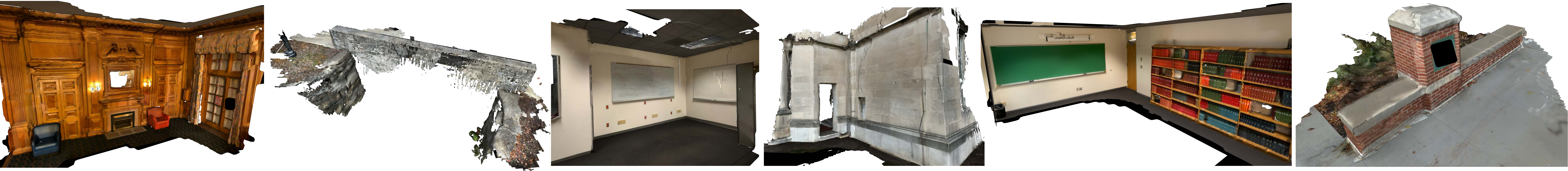}
    \captionsetup{width=\linewidth}
    \caption{\textbf{Remaining six scenes with our textures}. See Appendix
    for all methods' results on these scenes.}
    \label{fig: qualitative more scenes}
\end{minipage}%

\begin{minipage}[t]{\columnwidth}
\centering
\begin{minipage}[t]{\columnwidth}
    \centering
    \includegraphics[width=0.7\textwidth]{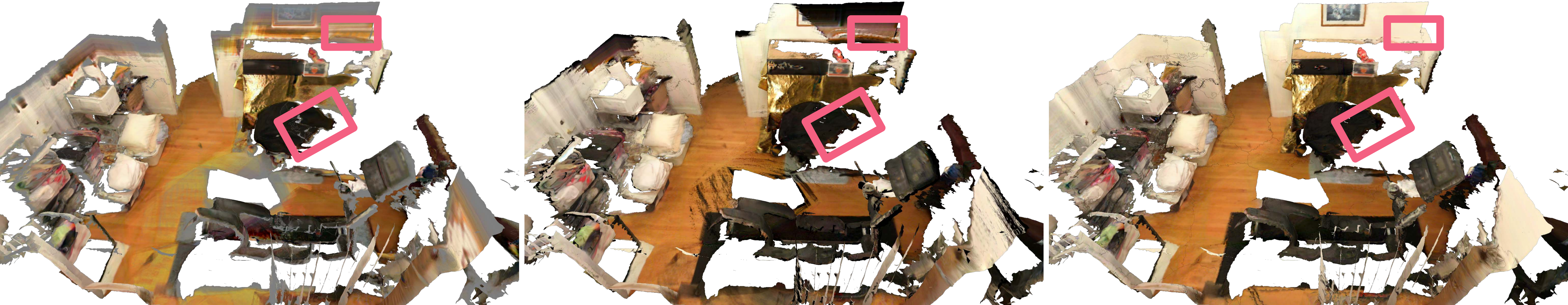}
    \captionsetup{width=\textwidth}
    \caption{\textbf{Results on ScanNet.} Left to right: AdvTex-B/C and ours.
        Ours alleviates artifacts: colors from box on the cabinet are mapped to the backpack and wall.
    }
    \label{fig: scannet}
\end{minipage}%
\end{minipage}
\end{figure}

\noindent \textbf{Qualitative evaluation.}
We present qualitative examples in~\figref{fig: qualitative}.
\figref{fig: qualitative L2Avg} and~\figref{fig: qualitative ColorMap} demonstrate that L2Avg and ColorMap produce overly smooth texture.
Meanwhile, due to noise in the geometry,~\eg,~\figref{fig: scene complexity}, TexMap fails to resolve texture seams and cannot produce a complete texture (\figref{fig: qualitative TexMap}).
MVSTex results in~\figref{fig: qualitative mvstex} are undesirable as geometries are removed.
This is because MVSTex requires ray collision checking to remove occluded faces. Due to the misalignment between geometries and cameras,  artifacts are introduced.
We show results of AdvTex-C in \figref{fig: qualitative AdvTex} as it outperforms AdvTex-B from~\tabref{tab: qunatitative}. Artifacts are highlighted.
Our method can largly mitigate such seams, which can be inferred from~\figref{fig: qualitative ours}.
In~\figref{fig: qualitative highlights}, we show renderings, which demonstrate the effectiveness of the proposed method.
Please see Appendix
for complete qualitative results of scenes in~\figref{fig: qualitative more scenes}.\\

\noindent\textbf{On ScanNet~\cite{Dai2017ScanNetR3}.}
Following~\cite{Huang2020AdversarialTO}, we study scenes with ID $\leq$ 20 ( \figref{fig: scannet},~\tabref{tab: qunatitative scannet}).
We improve upon baselines (AdvTex-B/C) by a margin on SSIM (0.534 $\rightarrow$ 0.571), LPIPS (0.557 $\rightarrow$ 0.503), $S_3$ (0.143 $\rightarrow$ 0.127), and Grad (3.753 $\rightarrow$ 3.324).

\csection{Conclusion}
We develop an initialization and an alignment method for fully-automatic texture generation from a given scene mesh, and a given sequence of RGBD images and their camera parameters. %
We observe the proposed method to yield appealing results, addressing robustness issues due to noisy geometry and misalignment of prior work. Quantitatively we observe improvements on both perceptual similarity (LPIPS from 0.335 to 0.309) and sharpness ($S_3$ from 0.135 to 0.120).

$\newline$
\noindent\textbf{Acknowledgements:} Supported in part by NSF grants 1718221, 2008387, 2045586, 2106825, MRI \#1725729, and NIFA award 2020-67021-32799.

\clearpage
{\small
\bibliographystyle{splncs04}
\bibliography{egbib}
}

\clearpage
\beginsupplement
\appendix

\section*{\Large\centering Supplementary Material: \\Initialization and Alignment \\for Adversarial Texture Optimization}

\renewcommand{\thesection}{\Alph{section}}

This supplementary is structured as follows:
\begin{itemize}
    \item \secref{supp sec: stats} provides details regarding the scenes;
    \item \secref{supp sec: imp details} describes  implementation details;
    \item \secref{supp sec: quant} provides more quantitative results;
    \item \secref{supp sec: more ablations} gives more ablation studies;
    \item \secref{supp sec: qualitative} shows more qualitative results.
\end{itemize}

\section{UofI Texture Scenes Details}\label{supp sec: stats}

We provide detailed scene statistics in~\tabref{supp tab: scene stats}. As mentioned in~\secref{sec: exp setup}, we collect a dataset of 11 scenes: four indoor and seven outdoor scenes. This dataset consists of a total of 2807 frames, of which 91, 2052, and 664 are of resolution $480\times 360$, $960\times720$, and $1920\times 1440$ respectively. For each scene, we reserve 10\% of the views for evaluation. In contrast, prior work~\cite{Huang2020AdversarialTO} ``select(s) 10 views uniformly distributed from the scanning video'' for evaluation while using up to thousands of frames  for texture generation. The studied setting is hence more demanding.

\section{Implementation Details}\label{supp sec: imp details}

\subsection{Network Structure for \texttt{TexSmooth}}

As mentioned in~\secref{sec: tex smooth}, we utilize a convolutional neural network for the discriminator $D$ in~\equref{eq: adv optim loss}. Specifically, we follow~\cite{Huang2020AdversarialTO}'s code release\footnote{\label{supp footnote:advtex equ}\url{https://github.com/hjwdzh/AdversarialTexture}} to utilize five convolutional layers with structures (6, 64, 2), (64, 128, 2), (128, 128, 2), (128, 128, 1), (128, 1, 1), where $(in, out, s)$ indicates the number of input channels, the number of output channels, and stride respectively. All layers employ a $4\times4$ kernel and use \texttt{VALID} padding. Regarding activation functions, the first four layers contain leakyReLUs while the last one uses a sigmoid activation.

\subsection{Hyperparameters}
The weight for the $\mathcal{L}_1$ loss in~\equref{eq: adv optim loss} is $\lambda = 10$. We decay it by a factor of 0.8 every 960 steps. We use Adam~\cite{Kingma2015AdamAM} with $\beta_1 = 0.5$ and $\beta_2 = 0.999$. The learning rates for the texture and discriminator are set to  $1e^{-3}$ and $1e^{-4}$ respectively. We run \texttt{AdvOptim} for 4000 iterations for AdvTex-C and our \texttt{TexSmooth} stage, following the code release.\cref{supp footnote:advtex equ} For \texttt{AdvTex-B}, \texttt{AdvOptim} runs for 50000 iterations as stated in their paper~\cite{Huang2020AdversarialTO}.

\subsection{\texttt{AdvOptim} Reimplementation}
\vspace{-0.1cm}

As mentioned in~\secref{sec: exp setup}, we re-implement \texttt{AdvOptim} using PyTorch based on the official TensorFlow (TF) code. To verify the correctness of our implementation, we compare results from the official code release and our re-implemented version on the Chair dataset from~\cite{Huang2020AdversarialTO}, which contains 35 scanned chairs. Specifically,
similar to~\secref{sec: exp setup}, we reserve 10\% of the views from each scan for evaluation, resulting in 14,836 training views and 1663 test views. As can be seen from the  $1^\text{st}$~\vs~$2^\text{nd}$ row in~\tabref{supp tab: chair}, we observe 0.830~\vs~0.828 SSIM, 0.163~\vs~0.167 LPIPS, 0.068~\vs~0.069 $S_3$, and 2.052~\vs~2.047 Grad. This verifies the correctness of our re-implementation.

\begin{table}[t]
\renewcommand{\arraystretch}{0.9}
\begin{adjustwidth}{0.0cm}{}
\captionsetup{width=\columnwidth}
\caption{\textbf{Statistics of UofI Texture Scenes.}
For each column, the format is indicated below the header.
We reserve 10\% views for evaluation. 
``Angular Diff'' measures the the angular differences between test set view directions and their nearest neighbour in the training sets.
}
\label{supp tab: scene stats}
\renewcommand\theadfont{}
\centering
{
\begin{tabular}{c|rrrr}
\toprule
& \makecell{Mesh\\\texttt{$\#$Faces}} & \makecell{$\#$Views\\\texttt{total (test)}} & \makecell{Resolution\\\texttt{value ($\#$count)}} & \makecell[c]{Angular Diff\\\texttt{avg (min/max)}} \\
\midrule
1 & 77,528 & 160 (16)  &  1920{$\times$}1440 (160) & 2.04$^\circ$ (0.38$^\circ$/4.26$^\circ$) \\
\midrule
2 & 104,684 & 195 (20)  &  960{$\times$}720 (195) & 0.85$^\circ$ (0.06$^\circ$/2.46$^\circ$) \\
\midrule
3 & 132,196 & 94 (10)  &  \makecell[r]{1920{$\times$}1440 (3)\\480{$\times$}360 (91)} & 4.58$^\circ$ (1.14$^\circ$/13.8$^\circ$)\\
\midrule
4 & 65,832 & 43 (5)  &  960{$\times$}720 (43) & 3.48$^\circ$ (0.12$^\circ$/6.66$^\circ$) \\
\midrule
5 & 143,108 & 584 (59)  &  960{$\times$}720 (584) & 1.28$^\circ$ (0.19$^\circ$/3.18$^\circ$)\\
\midrule
6 & 168,664 & 372 (38)  &  1920{$\times$}1440 (372) & 1.45$^\circ$ (0.36$^\circ$/3.13$^\circ$) \\
\midrule
7 & 77,627 & 233 (24)  &  960{$\times$}720 (233) & 3.24$^\circ$ (0.67$^\circ$/12.6$^\circ$) \\
\midrule
8 & 80,240 & 352 (36)  &  960{$\times$}720 (352) & 1.60$^\circ$ (0.27$^\circ$/6.89$^\circ$) \\
\midrule
9 & 199,976 & 347 (35)  &  960{$\times$}720 (347) & 1.81$^\circ$ (0.21$^\circ$/4.55$^\circ$) \\
\midrule
10 & 69,484 & 129 (13)  &  1920{$\times$}1440 (129) & 1.12$^\circ$ (0.19$^\circ$/2.59$^\circ$) \\
\midrule
11 & 149,176 & 298 (30)  &  960{$\times$}720 (298) & 1.13$^\circ$ (0.27$^\circ$/2.60$^\circ$) \\
\toprule
\end{tabular}
}
\end{adjustwidth}
\end{table}

\begin{table}[t]
\renewcommand{\arraystretch}{1}
\begin{adjustwidth}{0.0cm}{}
\captionsetup{width=\linewidth}
\caption{
\textbf{Quantitative results on Chair dataset}. We report in the form of $\texttt{mean}{\scriptsize\pm\texttt{std}}$.
}
\label{supp tab: chair}
\renewcommand\theadfont{}
\centering
{
\small
\begin{tabular}{l@{\hskip 0.4em}l@{\hskip 0.6em}c@{\hskip 0.4em}r@{\hskip 0.65em}r@{\hskip 0.65em}r} 
\toprule
 &&  SSIM$\uparrow$  & LPIPS$\downarrow$ & $S_3\downarrow$ & Grad$\downarrow$ \\
\midrule
1 & TF & 0.830{\scriptsize$\pm$0.072}  & 0.163{\scriptsize$\pm$0.072}  & 0.068{\scriptsize$\pm$0.020}  & 2.052{\scriptsize$\pm$1.267}  \\
2 & PyTorch & 0.828{\scriptsize$\pm$0.072}  & 0.167{\scriptsize$\pm$0.073}  & 0.069{\scriptsize$\pm$0.020}  & 2.047{\scriptsize$\pm$1.275}  \\
\midrule
3 & \makecell[l]{\texttt{TexInit} +\\ \texttt{TexSmooth}} & 0.827{\scriptsize$\pm$0.073}  & 0.167{\scriptsize$\pm$0.073}  & 0.068{\scriptsize$\pm$0.019}  & 2.071{\scriptsize$\pm$1.289} \\
\toprule
\end{tabular}
}
\end{adjustwidth}
\end{table}

\subsection{Determine $\cTfinal$ resolution}
To determine the $H$ and $W$ of the texture $\cTfinal$, we use the following three steps: 
1) for each of the $|\cY|$ planes mentioned in~\secref{sec: atlas gen}'s ``Overlap Detection'', we use a resolution of $512^2$, $1024^2$, or $2048^2$ based on whether the major RGB resolution's larger side  is 480, 960, or 1920;
2) all $|\cY|$ planes are concatenated to obtain the texture $\cTfinal$'s $H$ and $W$;
3) For a fair comparison, AdvTex baselines use the same resolution as ours.

\vspace{-0.1cm}
\section{More Quantitative Results}\label{supp sec: quant}

\subsection{Scene-level Quantitative Results}\label{supp sec: scene-level quant}

We provide scene-level quantitative results in~\tabref{supp tab: scene qunatitative}. As can be seen from the  number for the Best or $2^\text{nd}$-Best results (last two rows), our technique improves upon all  baselines.
Specifically, ours dominates the count (ours~\vs~runner-up) for ``best'' (17~\vs~14), ``2$^{\text{nd}}$-best'' (14~\vs~13), and ``best or 2$^{\text{nd}}$-best'' (31~\vs~18).

\subsection{Results on Existing Dataset}

We apply our \texttt{TexInit} + \texttt{TexSmooth}  on the Chair dataset. We directly utilize the provided conformal mapping to ensure a fair comparison. Quantitative results are shown in $2^\text{nd}$~\vs~$3^\text{rd}$ row of~\tabref{supp tab: chair}. Our method performs on par with the baseline. This is expected as we do not observe geometry and images of those scans to be misaligned. Beneficially and as expected, the proposed technique doesn't harm the result if geometry and images are well aligned. %

\subsection{Robustness to Inaccurate Camera Poses}

To verify the robustness of the proposed pipeline, we conduct studies by deliberately adding more noise to camera poses in the training split. Concretely, given a camera pose with rotation $(r_x, r_y, r_z)$ (represented in Euler angles) and translation $(t_x, t_y, t_z)$, we add uniformly-sampled noise, \ie, we have $\widehat{r}_x = r_x + \epsilon_{r_x}$, where $\epsilon_{r_x} \sim \mathcal{U}(-0.05 \cdot \vert r_x \vert, 0.05 \cdot \vert r_x \vert)$, and analogously $\widehat{r}_y, \widehat{r}_z, \widehat{t}_x, \widehat{t}_y, \widehat{t}_z$.
We apply AdvTex-B/C and ours on data with these corrupted camera poses.~\tabref{supp tab: cam noises} corroborates the robustness of the method: we outperform baselines on SSIM (0.456~\vs~0.452, $\uparrow$ is better), LPIPS (0.472~\vs~0.490, $\downarrow$ is better), and sharpness $S_3$ (0.142~\vs~0.153, $\downarrow$ is better).

\begin{table}[t]
\renewcommand{\arraystretch}{1.0}
\begin{adjustwidth}{0.0cm}{}
\captionsetup{width=\linewidth}
\caption{
\textbf{Evaluation with noise added to camera poses on UofI Texture Scenes}.
Results are in the form of $\texttt{mean}{\scriptsize\pm\texttt{std}}$. 
Ours is the most robust.
}
\label{supp tab: cam noises}
\renewcommand\theadfont{}
\centering
\setlength{\tabcolsep}{3pt}
{
\scriptsize
\begin{tabular}{lcrrrr} 
\toprule
 &  & SSIM$\uparrow$  & LPIPS$\downarrow$ & $S_3\downarrow$ & Grad$\downarrow$ \\

\midrule
1-1 & AdvTex-B & 0.378{\scriptsize$\pm$0.158}  & 0.503{\scriptsize$\pm$0.091}  & 0.160{\scriptsize$\pm$0.040}  & 9.193{\scriptsize$\pm$4.562}   \\
1-2 & AdvTex-C & 0.452{\scriptsize$\pm$0.191}  & 0.490{\scriptsize$\pm$0.082}  & 0.153{\scriptsize$\pm$0.072}  & \textbf{7.949}{\scriptsize$\pm$4.364}   \\
\midrule
\belowrulesepcolor{highlightRowColor} 
\rowcolor{highlightRowColor}
2 & Ours & \textbf{0.456}{\scriptsize$\pm$0.196}  & \textbf{0.472}{\scriptsize$\pm$0.078}  & \textbf{0.142}{\scriptsize$\pm$0.050}  & 8.177{\scriptsize$\pm$4.441}    \\
\toprule
\end{tabular}
}
\end{adjustwidth}
\end{table}

\section{More Ablations}\label{supp sec: more ablations}
\vspace{-0.1cm}

\subsection{Unary~\vs~Pairwise Cues}
\vspace{-0.1cm}

As stated in~\equref{eq:MRFTex}, we use pure-unary cues for \texttt{TexInit}. To verify this design choice, we ablate with a setup where \texttt{TexInit} considers both unary and pairwise cues. Concretely, we consider the following optimization problem:
\begin{equation}
    \bt^\ast = \argmax_\bt \sum_{i=1}^{|M|} \psi_i(t_i) + \sum_{(i,j)\in\cA} \psi_{i,j}(t_i,t_j),
    \label{supp eq: MRFTex pairwise}
\end{equation}
where $\cA$ is the adjacency used in \equref{eq:MRFPlane}. 
Here $\psi_i$ refers to the \textit{unary} cues in~\equref{eq:MRFTex} while $\psi_{i,j}$ captures \textit{pairwise} ones. 
Therefore, besides $\cue{1}$, $\cue{2}$, and $\cue{3}$ discussed in~\secref{sec: atlas gen}, we take into account another cue $\cue{4}$:
\begin{equation}
  \psi_{i,j}(t_i,t_j) \doteq \psi_{i,j}^{\cue{4}}(t_i,t_j).
\end{equation}

\noindent $\bullet$~\textbf{Explicit adjacency encouragement} ($\cue{4}$). Intuitively, adjacent triangles $\texttt{Tri}_i$ and $\texttt{Tri}_j$ maintain  smoothness if they are assigned textures from the same frame. We  encourage this choice using $\psi_{i,j}^{\cue{4}}(t_i,t_j) = \mathbbm{1}(t_i = t_j)$ and set $\omega_4 > 0$.

In our experiments, we set $\omega_4 = 1$ for \texttt{TexInit}. The \texttt{TexSmooth} stage remains the same. Quantitative results are shown in Column 1-1~\vs~1-2 in~\tabref{tab: scene sensitivity + pairwise}. We do not observe a big difference when integrating pairwise cues. Concretely, when comparing pairwise~\vs~unary only cues, we have 0.601~\vs~0.602 (SSIM), 0.305~\vs~0.309 (LPIPS), 0.120~\vs~0.120 ($S_3$), and 6.872~\vs~6.871 (Grad), which corroborate our design.

\subsection{View Sparsity Analysis}

To understand whether our framework is sensitive to the sparsity of training views, we conduct an ablation: 1) we reserve 10\% of the views, which are uniformly sampled, for evaluation. As a consequence, we have 90\% of all views that can be used for training; 2) within those 90\% of all views, we again uniformly sample images for optimization every $k$ views. We use $k \in \{1, 2, 3, 4, 5\}$. Note, the results reported in~\tabref{tab: qunatitative},~\tabref{tab: ablation}, and~\tabref{supp tab: scene qunatitative} are situations where $k=1$, namely all 90\% views are used. Quantitative results are reported in Column 1-1 to Column 5 in~\tabref{tab: scene sensitivity + pairwise}. As expected, when the number of training views decreases, we observe the results' quality to drop. However, our framework is robust to the view sparsity as the performance gap is small. Specifically, we observe best~\vs~worst results as 0.604~\vs~0.580 (SSIM), 0.303~\vs~0.316 (LPIPS), 0.120~\vs~0.127 ($S_3$), and 6.859~\vs~7.026 (Grad). The reason that $k=2$ performs slightly better than $k=1$ is that 1) $k=2$ still provides dense enough views for \texttt{TexInit} and \texttt{TexSmooth}; 2) the initialization from \texttt{TexInit} can contain less seams as less views are considered.

\subsection{Patchwise Alignment}

\begin{figure}[t]
\centering
    \centering
    \includegraphics[width=0.7\columnwidth]{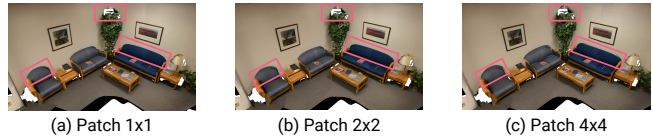}
    \caption{\textbf{Patch-wise alignment on UofI Texture Scenes}. Highlighted issues: sofa and plant colors are mapped to the wall in (b,c).} 
    \label{supp fig: patch}
\end{figure}

We assess the results of  patch-wise alignment in our pipeline.~\tabref{supp tab: patch} and~\figref{supp fig: patch} verify: a patch-wise method ignores global content and is inferior.

\begin{table}[t]
\renewcommand{\arraystretch}{1.0}
\begin{adjustwidth}{0.0cm}{}
\captionsetup{width=\linewidth}
\caption{
\textbf{Patchwise alignment on UofI Texture Scenes}.
Results are in the form of $\texttt{mean}{\scriptsize\pm\texttt{std}}$. 
 $X\times Y$ in the `\#Patches' column denotes the number of patches created by splitting  along height ($X$) and width ($Y$). See~\figref{supp fig: patch} for qualitative results.
}
\label{supp tab: patch}
\renewcommand\theadfont{}
\centering
\setlength{\tabcolsep}{3pt}
{
\scriptsize
\begin{tabular}{lcrrrr} 
\toprule
 & \#Patches & SSIM$\uparrow$  & LPIPS$\downarrow$ & $S_3\downarrow$ & Grad$\downarrow$ \\

\midrule
1 & $1\times1$ & \textbf{0.602}{\scriptsize$\pm$0.189}  & \textbf{0.309}{\scriptsize$\pm$0.086}  & \textbf{0.120}{\scriptsize$\pm$0.058}  & \textbf{6.871}{\scriptsize$\pm$4.342}   \\
2 & $2\times2$ & 0.556{\scriptsize$\pm$0.184}  & 0.330{\scriptsize$\pm$0.069}  & 0.123{\scriptsize$\pm$0.055}  & 7.122{\scriptsize$\pm$4.306}   \\
3 & $4\times4$ & 0.545{\scriptsize$\pm$0.175}  & 0.338{\scriptsize$\pm$0.064}  & 0.122{\scriptsize$\pm$0.054}  & 7.163{\scriptsize$\pm$4.257}   \\
\toprule
\end{tabular}
}
\end{adjustwidth}
\end{table}

\begin{table*}[t]

\captionsetup{width=\textwidth}
\caption{
\textbf{Scene-level quantitative results on UofI Texture Scenes.} We use \textbf{bold} and \underline{underline} to mark best and $2^\text{nd}$-best results for each row respectively. If one value's difference from its higher-rank counterpart is no larger than 0.001, we treat them as the same.
}
\label{supp tab: scene qunatitative}

\renewcommand{\arraystretch}{1.1}
\centering
\begin{adjustbox}{width=0.95\columnwidth,center}

\begin{tabular}{
l@{\hskip 0.8em}l g r g | r g r | g r g | r
} 

\specialrule{.15em}{.05em}{.05em}
\rowcolor{white}
 && \makecell[c]{1} & \makecell[c]{2} & \makecell[c]{3} & \makecell[c]{4} & \makecell[c]{5} & \makecell[c]{6} & \makecell[c]{7} & \makecell[c]{8} & \makecell[c]{9} & \makecell[c]{10}  \\
\rowcolor{white}
&& \makecell[c]{L2Avg} & \makecell[c]{ColorMap} & \makecell[c]{TexMap} & MVSTex & \makecell[c]{AdvTex-B} & \makecell[c]{AdvTex-C} & \makecell[c]{$\cT$ Only} & \makecell[c]{w/o $\cT$} & \makecell[c]{w/o \texttt{Align}} & \makecell[c]{Ours} \\
\midrule
\parbox[t]{2mm}{\multirow{4}{*}{\rotatebox[origin=c]{90}{\small\makecell{Scene 1}}}} & SSIM$\uparrow$ & \underline{0.866}  & 0.831  & 0.608  & 0.718  & 0.774  & 0.834  & 0.769  & 0.863  & 0.845  & \textbf{0.876}  \\
& LPIPS$\downarrow$ & 0.235  & 0.361  & 0.387  & 0.260  & 0.227  & 0.261  & 0.251  & \underline{0.215}  & 0.228  & \textbf{0.187} \\ 
& $S_3\downarrow$ & \textbf{0.042}  & 0.058  & 0.088  & 0.073  & 0.064  & 0.051  & 0.070  & 0.043  & 0.046  & \textbf{0.041} \\ 
& Grad$\downarrow$  & \underline{1.806}  & 2.085  & 3.232  & 2.977  & 2.349  & 2.073  & 2.414  & 1.816  & 1.936  & \textbf{1.724} \\ 
\midrule
\parbox[t]{2mm}{\multirow{4}{*}{\rotatebox[origin=c]{90}{\small\makecell{Scene 2}}}} & SSIM$\uparrow$ & 0.584  & 0.428  & 0.384  & 0.494  & 0.486  & 0.541  & 0.521  & \underline{0.592}  & 0.532  & \textbf{0.626}  \\
& LPIPS$\downarrow$ & 0.319  & 0.653  & 0.433  & \textbf{0.221}  & 0.288  & 0.290  & 0.271  & 0.231  & 0.294  & \underline{0.226} \\ 
& $S_3\downarrow$ & 0.227  & 0.342  & 0.253  & 0.172  & 0.172  & 0.180  & 0.190  & \textbf{0.165}  & 0.174  & \textbf{0.164} \\ 
& Grad$\downarrow$ & 8.910  & 11.09  & 11.75  & 10.28  & 9.909  & 9.499  & 9.749  & \underline{8.488}  & 9.707  & \textbf{8.067} \\ 
\midrule
\parbox[t]{2mm}{\multirow{4}{*}{\rotatebox[origin=c]{90}{\small\makecell{Scene 3}}}} & SSIM$\uparrow$ & \textbf{0.651}  & 0.594  & 0.465  & 0.575  & 0.534  & 0.605  & 0.561  & 0.630  & 0.585  & \underline{0.636}   \\
& LPIPS$\downarrow$  & 0.362  & 0.533  & 0.459  & \textbf{0.301}  & 0.383  & 0.344  & 0.354  & 0.324  & 0.348  & \underline{0.309}  \\ 
& $S_3\downarrow$ & 0.135  & 0.170  & 0.188  & 0.129  & 0.153  & \textbf{0.121}  & 0.132  & 0.128  & 0.124  & \textbf{0.121}  \\ 
& Grad$\downarrow$ & 5.721  & \underline{5.707}  & 8.203  & 6.836  & 7.468  & 6.051  & 6.566  & 6.118  & 6.238  & \textbf{5.659}  \\ 
\midrule
\parbox[t]{2mm}{\multirow{4}{*}{\rotatebox[origin=c]{90}{\small\makecell{Scene 4}}}} & SSIM$\uparrow$ & \underline{0.217}  & 0.182  & 0.168  & 0.155  & 0.133  & 0.187  & 0.176  & 0.198  & 0.180  & \textbf{0.244}  \\
& LPIPS$\downarrow$  & 0.656  & 0.845  & 0.468  & \textbf{0.370}  & 0.575  & 0.540  & \underline{0.430}  & 0.516  & 0.500  & 0.440 \\ 
& $S_3\downarrow$ & 0.310  & 0.441  & 0.168  & \textbf{0.112}  & 0.152  & 0.143  & \underline{0.118}  & 0.141  & 0.128  & 0.121 \\ 
& Grad$\downarrow$ & 16.79  & 18.35  & 16.68  & \textbf{15.78}  & 18.05  & 16.07  & 16.76  & \underline{15.95}  & 16.13  & 16.09  \\ 
\midrule
\parbox[t]{2mm}{\multirow{4}{*}{\rotatebox[origin=c]{90}{\small\makecell{Scene 5}}}} & SSIM$\uparrow$ & \textbf{0.480}  & 0.430  & 0.397  & 0.412  & 0.422  & 0.448  & 0.422  & \underline{0.474}  & 0.430  & 0.472 \\
& LPIPS$\downarrow$  & 0.427  & 0.615  & 0.391  & \textbf{0.268}  & 0.343  & 0.351  & 0.308  & 0.332  & 0.321  & \underline{0.293}  \\ 
& $S_3\downarrow$ & 0.326  & 0.403  & 0.267  & 0.213  & 0.225  & 0.227  & 0.223  & 0.220  & \underline{0.210}  & \textbf{0.205} \\ 
& Grad$\downarrow$ & 10.38  & 11.80  & 11.83  & 10.62  & 10.85  & \underline{9.359}  & 11.18  & \textbf{9.230}  & 10.12  & 9.807 \\ 
\midrule
\parbox[t]{2mm}{\multirow{4}{*}{\rotatebox[origin=c]{90}{\small\makecell{Scene 6}}}} & SSIM$\uparrow$ & \textbf{0.684}  & 0.615  & 0.330  & 0.392  & 0.509  & 0.598  & 0.480  & 0.637  & 0.591  & \underline{0.642}  \\
& LPIPS$\downarrow$ & 0.374  & 0.615  & 0.674  & 0.553  & 0.386  & 0.403  & 0.419  & \textbf{0.343}  & 0.408  & \underline{0.363}  \\ 
& $S_3\downarrow$ & \textbf{0.058}  & 0.099  & 0.122  & 0.109  & 0.129  & 0.066  & 0.127  & \underline{0.064}  & 0.075  & 0.067 \\ 
& Grad$\downarrow$ & \textbf{3.020}  & 3.552  & 4.925  & 4.790  & 4.500  & 3.435  & 4.476  & 3.360  & 3.551  & \underline{3.256} \\ 
\midrule
\parbox[t]{2mm}{\multirow{4}{*}{\rotatebox[origin=c]{90}{\small\makecell{Scene 7}}}} & SSIM$\uparrow$ & \textbf{0.423}  & 0.378  & 0.343  & 0.346  & 0.362  & 0.384  & 0.374  & \underline{0.411}  & 0.373  & 0.396  \\
& LPIPS$\downarrow$ & 0.465  & 0.650  & 0.475  & \textbf{0.347}  & 0.436  & 0.442  & \underline{0.356}  & 0.448  & 0.392  & 0.367 \\ 
& $S_3\downarrow$ & 0.280  & 0.385  & 0.262  & 0.215  & 0.216  & 0.245  & 0.220  & 0.239  & \underline{0.203}  & \textbf{0.200} \\ 
& Grad$\downarrow$ & 12.16  & 13.98  & 13.27  & 12.93  & 13.27  & 12.09  & 13.65  & \textbf{11.60}  & 12.11  & \underline{11.91}  \\
\midrule
\parbox[t]{2mm}{\multirow{4}{*}{\rotatebox[origin=c]{90}{\small\makecell{Scene 8}}}} & SSIM$\uparrow$ & \textbf{0.846}  & 0.808  & 0.265  & 0.629  & 0.679  & 0.815  & 0.740  & 0.839  & 0.816  & \underline{0.842} \\
& LPIPS$\downarrow$  & 0.269  & 0.498  & 0.680  & 0.396  & 0.328  & 0.286  & 0.349  & \underline{0.259}  & 0.278  & \textbf{0.250} \\ 
& $S_3\downarrow$ & 0.060  & 0.084  & 0.135  & 0.117  & 0.096  & 0.063  & 0.080  & \underline{0.058}  & 0.060  & \textbf{0.055}  \\ 
& Grad$\downarrow$ & \underline{2.062}  & 2.489  & 4.627  & 4.405  & 3.181  & 2.275  & 2.877  & 2.070  & 2.257  & \textbf{2.011} \\ 
\midrule
\parbox[t]{2mm}{\multirow{4}{*}{\rotatebox[origin=c]{90}{\small\makecell{Scene 9}}}} & SSIM$\uparrow$ & \textbf{0.545}  & 0.502  & 0.306  & 0.373  & 0.376  & 0.450  & 0.383  & 0.500  & 0.469  & \underline{0.508} \\
& LPIPS$\downarrow$ & 0.483  & 0.709  & 0.509  & \textbf{0.371}  & 0.460  & 0.520  & \underline{0.419}  & 0.474  & 0.454  & 0.451  \\ 
& $S_3\downarrow$ & 0.194  & 0.254  & 0.197  & 0.146  & 0.146  & 0.163  & \textbf{0.132}  & 0.159  & \underline{0.138}  & 0.143  \\ 
& Grad$\downarrow$ & 6.988  & 8.042  & 9.815  & 8.751  & 8.582  & 7.514  & 8.380  & \textbf{6.765}  & 6.985  & \underline{6.814} \\ 
\midrule
\parbox[t]{2mm}{\multirow{4}{*}{\rotatebox[origin=c]{90}{\small\makecell{Scene 10}}}} & SSIM$\uparrow$ & \textbf{0.853}  & 0.800  & 0.493  & 0.718  & 0.716  & 0.810  & 0.744  & 0.828  & 0.816  & \underline{0.839} \\
& LPIPS$\downarrow$ & 0.260  & 0.403  & 0.492  & 0.305  & 0.282  & 0.244  & 0.267  & \underline{0.204}  & 0.240  & \textbf{0.198} \\ 
& $S_3\downarrow$ & \textbf{0.043}  & 0.059  & 0.088  & 0.072  & 0.099  & 0.051  & 0.081  & 0.047  & 0.049  & \underline{0.045} \\ 
& Grad$\downarrow$ & \textbf{2.159}  & 2.476  & 3.740  & 3.366  & 3.295  & 2.564  & 3.081  & 2.416  & 2.421  & \underline{2.283} \\ 
\midrule
\parbox[t]{2mm}{\multirow{4}{*}{\rotatebox[origin=c]{90}{\small\makecell{Scene 11}}}} & SSIM$\uparrow$ & \textbf{0.563}  & 0.520  & 0.374  & 0.428  & 0.454  & 0.518  & 0.443  & \underline{0.546}  & 0.513  & 0.544  \\
& LPIPS$\downarrow$ & 0.398  & 0.510  & 0.399  & \textbf{0.292}  & 0.350  & 0.337  & 0.334  & \underline{0.307}  & 0.345  & 0.312 \\ 
& $S_3\downarrow$ & 0.229  & 0.278  & 0.198  & 0.167  & 0.178  & 0.177  & 0.179  & \underline{0.165}  & 0.167  & \textbf{0.158} \\ 
& Grad$\downarrow$ & \textbf{7.742}  & 8.100  & 10.04  & 9.435  & 9.057  & \underline{7.946}  & 9.878  & 7.684  & 8.225  & 7.957 \\ 
\toprule
\parbox[t]{2mm}{\multirow{2}{*}{\rotatebox[origin=c]{90}{\small\makecell{Count}}}} & Best & 14 & 0 & 0 & 9 & 0 & 1 & 1 & 5 & 0 & \textbf{17} \\
& $2^\text{nd}$-Best & 4 & 1 & 0 & 0 & 0 & 2 & 4 & 13 & 3 & \textbf{14} \\
\specialrule{.15em}{.05em}{.05em}
\end{tabular}
\end{adjustbox}
\end{table*}

\begin{table*}[t]

\captionsetup{width=\textwidth}
\caption{
\textbf{Scene-level ablation results on UofI Texture Scenes.}
The fractions below each header indicate the portion of training views utilized during optimization. Note, we do not consider the already-reserved 10\% views for evaluation. Namely, $1/1$ means we use all 90\% training views.
For Column 1-1, we use pairwise cues for \texttt{TexInit} while only unary cues are utilized for the remaining columns.
We report $\texttt{mean}{\scriptsize\pm\texttt{std}}$.
}
\label{tab: scene sensitivity + pairwise}

\renewcommand{\arraystretch}{1.1}
\centering
\resizebox*{!}{0.9\textheight}{
{
\setlength{\tabcolsep}{5pt}

\begin{tabular}{
l@{\hskip 0.8em}l g | r g r g r
} 

\specialrule{.15em}{.05em}{.05em}
\rowcolor{white}
 && \makecell[c]{1-1} & \makecell[c]{1-2} & \makecell[c]{2} & \makecell[c]{3} & \makecell[c]{4} & \makecell[c]{5}  \\
\rowcolor{white}
&& \makecell[c]{1/1-pairwise} & \makecell[c]{1/1} & \makecell[c]{1/2} & \makecell[c]{1/3} & \makecell[c]{1/4} & \makecell[c]{1/5} \\
\midrule
\parbox[t]{2mm}{\multirow{4}{*}{\rotatebox[origin=c]{90}{\small\makecell{Scene 1}}}} & SSIM$\uparrow$ & 0.876  & 0.876  & 0.873  & 0.868  & 0.831  & 0.826   \\
& LPIPS$\downarrow$ & 0.186  & 0.187  & 0.185  & 0.187  & 0.216  & 0.213 \\ 
& $S_3\downarrow$ & 0.040  & 0.041  & 0.042  & 0.044  & 0.058  & 0.063 \\ 
& Grad$\downarrow$ & 1.727  & 1.724  & 1.742  & 1.770  & 2.045  & 2.120 \\ 
\midrule
\parbox[t]{2mm}{\multirow{4}{*}{\rotatebox[origin=c]{90}{\small\makecell{Scene 2}}}} & SSIM$\uparrow$ & 0.622  & 0.626  & 0.624  & 0.617  & 0.614  & 0.610   \\
& LPIPS$\downarrow$ & 0.219  & 0.226  & 0.218  & 0.215  & 0.221  & 0.215 \\ 
& $S_3\downarrow$ & 0.162  & 0.164  & 0.160  & 0.161  & 0.161  & 0.159 \\ 
& Grad$\downarrow$ & 8.070  & 8.067  & 8.053  & 8.105  & 8.162  & 8.124 \\ 
\midrule
\parbox[t]{2mm}{\multirow{4}{*}{\rotatebox[origin=c]{90}{\small\makecell{Scene 3}}}} & SSIM$\uparrow$ & 0.637  & 0.636  & 0.657  & 0.630  & 0.601  & 0.639 \\
& LPIPS$\downarrow$ & 0.307  & 0.309  & 0.293  & 0.312  & 0.314  & 0.265 \\ 
& $S_3\downarrow$ & 0.120  & 0.121  & 0.117  & 0.125  & 0.133  & 0.112 \\ 
& Grad$\downarrow$ & 5.639  & 5.659  & 5.445  & 5.706  & 6.008  & 5.435 \\  
\midrule
\parbox[t]{2mm}{\multirow{4}{*}{\rotatebox[origin=c]{90}{\small\makecell{Scene 4}}}} & SSIM$\uparrow$ & 0.237  & 0.244  & 0.258  & 0.240  & 0.219  & 0.204   \\
& LPIPS$\downarrow$ & 0.442  & 0.440  & 0.440  & 0.444  & 0.466  & 0.417 \\ 
& $S_3\downarrow$ & 0.123  & 0.121  & 0.122  & 0.128  & 0.128  & 0.114 \\ 
& Grad$\downarrow$ & 16.21  & 16.09  & 16.09  & 16.33  & 16.00  & 16.45  \\ 
\midrule
\parbox[t]{2mm}{\multirow{4}{*}{\rotatebox[origin=c]{90}{\small\makecell{Scene 5}}}} & SSIM$\uparrow$ & 0.473  & 0.472  & 0.471  & 0.473  & 0.473  & 0.474  \\
& LPIPS$\downarrow$ & 0.294  & 0.293  & 0.288  & 0.288  & 0.290  & 0.287  \\ 
& $S_3\downarrow$ & 0.206 & 0.205  & 0.205  & 0.205  & 0.205  & 0.203 \\ 
& Grad$\downarrow$ & 9.853  & 9.807  & 9.866  & 9.830  & 9.850  & 9.889  \\ 
\midrule
\parbox[t]{2mm}{\multirow{4}{*}{\rotatebox[origin=c]{90}{\small\makecell{Scene 6}}}} & SSIM$\uparrow$ & 0.644  & 0.642  & 0.643  & 0.627  & 0.621  & 0.618  \\
& LPIPS$\downarrow$ & 0.364  & 0.363  & 0.351  & 0.351  & 0.355  & 0.353  \\ 
& $S_3\downarrow$ & 0.066  & 0.067  & 0.070  & 0.073  & 0.074  & 0.073 \\ 
& Grad$\downarrow$ & 3.245  & 3.256  & 3.284  & 3.349  & 3.375  & 3.352  \\ 
\midrule
\parbox[t]{2mm}{\multirow{4}{*}{\rotatebox[origin=c]{90}{\small\makecell{Scene 7}}}} & SSIM$\uparrow$ & 0.400  & 0.396  & 0.407  & 0.400  & 0.400  & 0.411   \\
& LPIPS$\downarrow$ & 0.367  & 0.367  & 0.368  & 0.366  & 0.366  & 0.354  \\ 
& $S_3\downarrow$ & 0.201  & 0.200  & 0.199  & 0.200  & 0.201  & 0.194  \\ 
& Grad$\downarrow$ & 11.88  & 11.91  & 11.74  & 11.94  & 11.93  & 11.62 \\ 
\midrule
\parbox[t]{2mm}{\multirow{4}{*}{\rotatebox[origin=c]{90}{\small\makecell{Scene 8}}}} & SSIM$\uparrow$ & 0.837  & 0.842  & 0.827  & 0.816  & 0.814  & 0.815 \\
& LPIPS$\downarrow$ & 0.249  & 0.250  & 0.257  & 0.271  & 0.270  & 0.268 \\ 
& $S_3\downarrow$ & 0.055  & 0.055  & 0.060  & 0.065  & 0.065  & 0.065 \\ 
& Grad$\downarrow$ & 2.035  & 2.011  & 2.115  & 2.209  & 2.224  & 2.224 \\ 
\midrule
\parbox[t]{2mm}{\multirow{4}{*}{\rotatebox[origin=c]{90}{\small\makecell{Scene 9}}}} & SSIM$\uparrow$ & 0.509  & 0.508  & 0.506  & 0.498  & 0.490  & 0.489  \\
& LPIPS$\downarrow$ & 0.440  & 0.451  & 0.441  & 0.441  & 0.441  & 0.428 \\ 
& $S_3\downarrow$ & 0.143  & 0.143  & 0.142  & 0.141  & 0.142  & 0.140 \\ 
& Grad$\downarrow$ & 6.788  & 6.814  & 6.787  & 6.844  & 6.886  & 6.852 \\ 
\midrule
\parbox[t]{2mm}{\multirow{4}{*}{\rotatebox[origin=c]{90}{\small\makecell{Scene 10}}}} & SSIM$\uparrow$ & 0.839  & 0.839  & 0.839  & 0.826  & 0.802  & 0.801  \\
& LPIPS$\downarrow$ & 0.194  & 0.198  & 0.189  & 0.202  & 0.221  & 0.219 \\ 
& $S_3\downarrow$ & 0.044  & 0.045  & 0.046  & 0.051  & 0.067  & 0.069 \\ 
& Grad$\downarrow$ & 2.256  & 2.283  & 2.282  & 2.379  & 2.599  & 2.636 \\ 
\midrule
\parbox[t]{2mm}{\multirow{4}{*}{\rotatebox[origin=c]{90}{\small\makecell{Scene 11}}}} & SSIM$\uparrow$ & 0.540  & 0.544  & 0.537  & 0.530  & 0.518  & 0.523 \\
& LPIPS$\downarrow$ & 0.297  & 0.312  & 0.310  & 0.311  & 0.315  & 0.311 \\ 
& $S_3\downarrow$ & 0.155  & 0.158  & 0.157  & 0.159  & 0.160  & 0.159 \\ 
& Grad$\downarrow$ & 7.890  & 7.957  & 8.034  & 8.085  & 8.204  & 8.106 \\ 
\toprule
\parbox[t]{2mm}{\multirow{4}{*}{\rotatebox[origin=c]{90}{\small\makecell{Aggregated}}}} & SSIM$\uparrow$ & 0.601{\scriptsize$\pm$0.189}  & 0.602{\scriptsize$\pm$0.189}  & 0.604{\scriptsize$\pm$0.184}  & 0.593{\scriptsize$\pm$0.184}  & 0.580{\scriptsize$\pm$0.180}  & 0.583{\scriptsize$\pm$0.182} \\
& LPIPS$\downarrow$ & 0.305{\scriptsize$\pm$0.086}  & 0.309{\scriptsize$\pm$0.086}  & 0.304{\scriptsize$\pm$0.086}  & 0.308{\scriptsize$\pm$0.084}  & 0.316{\scriptsize$\pm$0.082}  & 0.303{\scriptsize$\pm$0.074}  \\ 
& $S_3\downarrow$ & 0.120{\scriptsize$\pm$0.058}  & 0.120{\scriptsize$\pm$0.058}  & 0.120{\scriptsize$\pm$0.056}  & 0.123{\scriptsize$\pm$0.055}  & 0.127{\scriptsize$\pm$0.051}  & 0.123{\scriptsize$\pm$0.049} \\ 
& Grad$\downarrow$ & 6.872{\scriptsize$\pm$4.364}  & 6.871{\scriptsize$\pm$4.342}  & 6.859{\scriptsize$\pm$4.321}  & 6.959{\scriptsize$\pm$4.355}  & 7.026{\scriptsize$\pm$4.233}  & 6.983{\scriptsize$\pm$4.296}  \\ 
\specialrule{.15em}{.05em}{.05em}
\end{tabular}
}
}
\end{table*}

\section{More Qualitative Results}\label{supp sec: qualitative}

\subsection{Alignment Visualizations}

\begin{figure}[t]
    \centering
    \includegraphics[height=0.9\textheight]{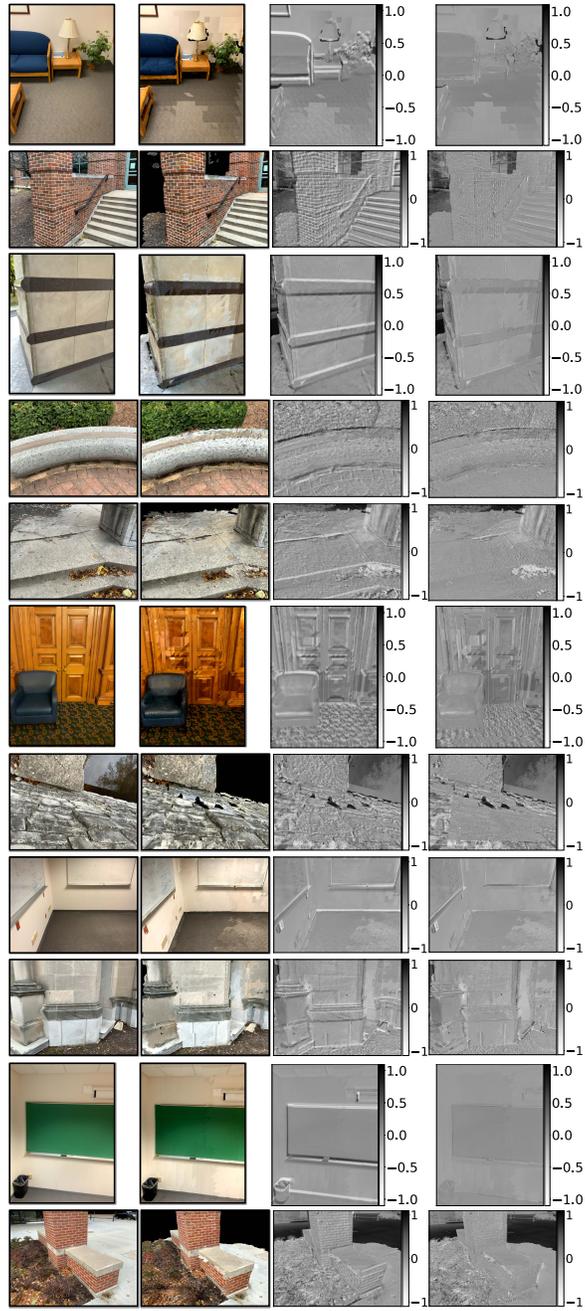}
    \vspace{-0.2cm}
    \caption{\textbf{Effect of alignment module.} From top to bottom, we show alignment results for Scene 1 to 11 respectively. For each scene, from left to right, we display the ground-truth image (GT), rendering from $\cT$, difference between GT and rendering before alignment, and difference after alignment.}
    \label{supp fig: offset}
    \vspace{-0.5cm}
\end{figure}

As mentioned in~\secref{sec: tex smooth}, we utilize the Fourier transformation to align ground-truth image and rendering from $\cT$.
We show qualitative results of the alignment for all 11 scenes in~\figref{supp fig: offset}. As can be seen clearly, our module successfully mitigates the misalignment, verifying the efficacy.

\subsection{Complete Qualitative Results}\label{supp sec: complete scenes}

\begin{figure*}[t]
\centering
    \vspace{0pt} 
    \captionsetup[subfigure]{aboveskip=1pt}
    \begin{subfigure}{\textwidth}
        \centering
        \includegraphics[width=\textwidth]{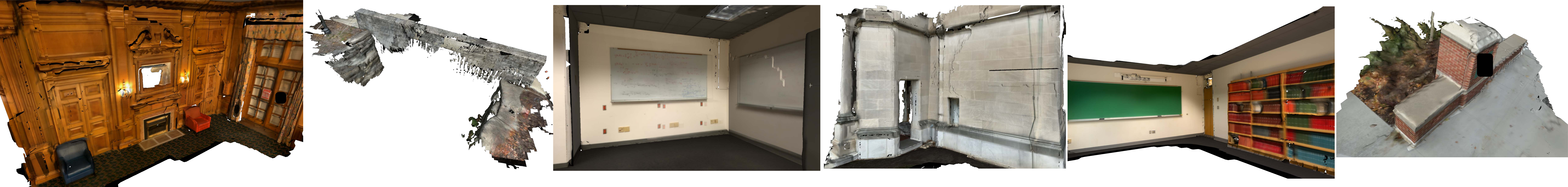}
        \captionsetup{width=\linewidth}
        \caption{\scriptsize\textbf{L2Avg.}}
        \label{supp fig: qualitative L2Avg}
    \end{subfigure}%
    \hfill
    \begin{subfigure}{\textwidth}
        \centering
        \includegraphics[width=\textwidth]{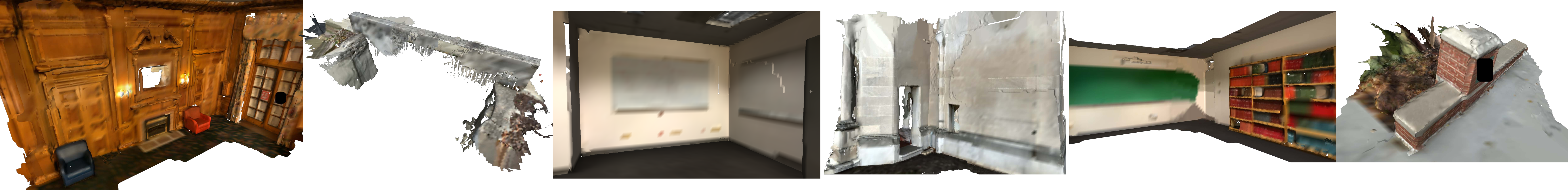}
        \captionsetup{width=\linewidth}
        \caption{\scriptsize\textbf{ColorMap}.}
        \label{supp fig: qualitative ColorMap}
    \end{subfigure}%
    \hfill
    \begin{subfigure}{\textwidth}
        \centering
        \includegraphics[width=\textwidth]{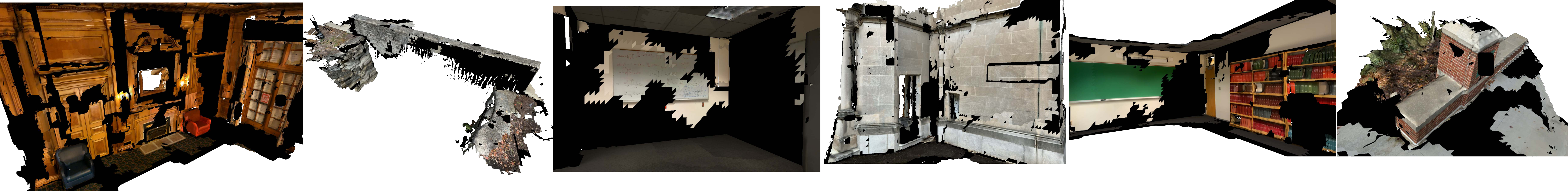}
        \captionsetup{width=\linewidth}
        \caption{\scriptsize\textbf{TexMap}.}
        \label{supp fig: qualitative TexMap}
    \end{subfigure}%
    \hfill
    \begin{subfigure}{\textwidth}
        \centering
        \includegraphics[width=\textwidth]{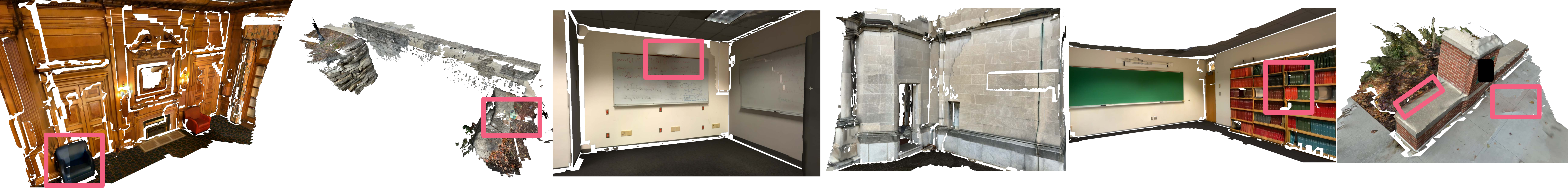}
        \captionsetup{width=\linewidth}
        \caption{\scriptsize\textbf{MVSTex}. Besides removing geometries, we also highlight artifacts that MVSTex produces.} 
        \label{supp fig: qualitative MVSTex}
    \end{subfigure}%
    \hfill
    \begin{subfigure}{\textwidth}
        \centering
        \includegraphics[width=\textwidth]{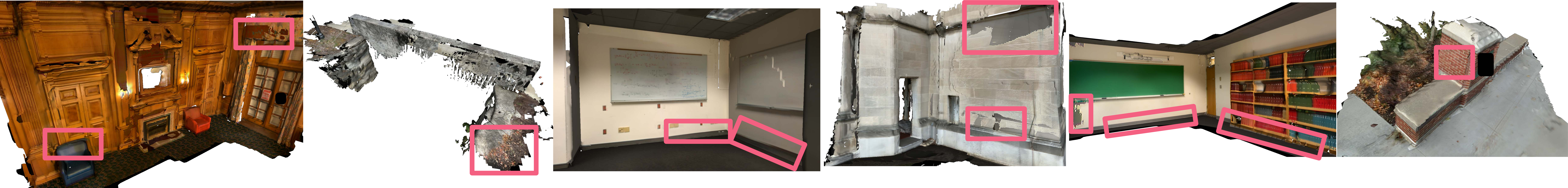}
        \captionsetup{width=\linewidth}
        \caption{\scriptsize\textbf{AdvTex-C}. We highlight artifacts:
        1) Scene 6: chair's texture is mapped to the door in the background and the curtain's pattern breaks;
        2) Scene 7: leaf colors are fused into the ground;
        3) Scene 8: the boundary between lower gray and upper white areas are fused;
        4) Scene 9: the color on the wall breaks;
        5) Scene 10: on the right: the texture of the bookshelf's lower part is mapped to the floor; on the left: the texture of carpet is mapped to the wall and the wall's color breaks;
        6) Scene 11: bricks' cracks are mixed together.
        }
        \label{supp fig: qualitative AdvTex}
    \end{subfigure}%
    \hfill
    \begin{subfigure}{\textwidth}
        \centering
        \includegraphics[width=\textwidth]{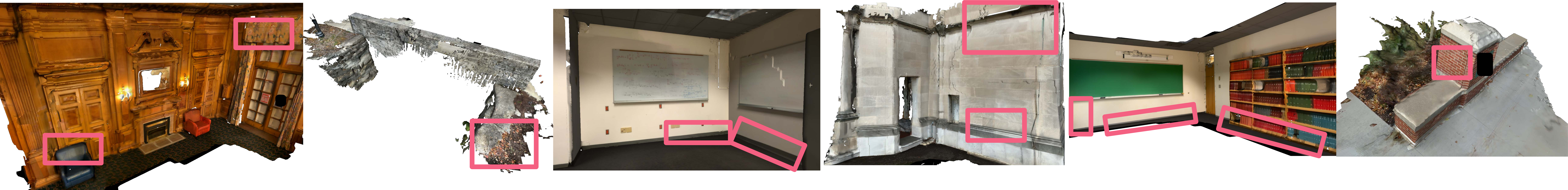}
        \captionsetup{width=\linewidth}
        \caption{\scriptsize\textbf{Ours}. Compared to~\figref{supp fig: qualitative AdvTex}, our method reduces artifacts.}
        \label{supp fig: qualitative ours}
    \end{subfigure}%
    \hfill
    \begin{subfigure}{\textwidth}
        \centering
        \includegraphics[width=\textwidth]{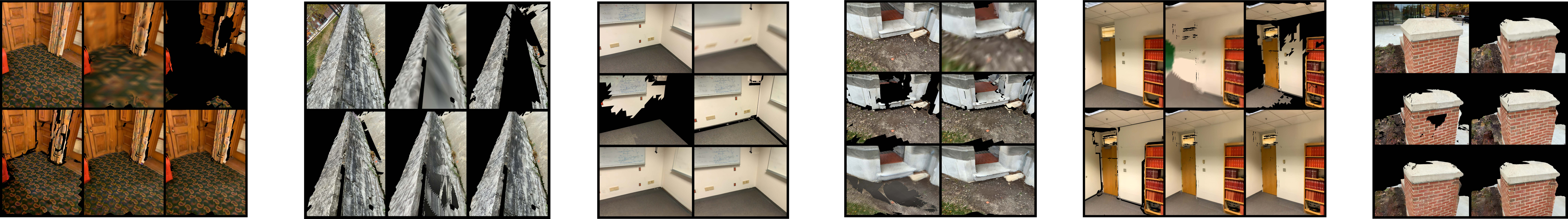}
        \captionsetup{width=\linewidth}
        \caption{\scriptsize\textbf{Highlights}. From left to right and top to bottom, we show ground-truth image as well as renderings at the same camera pose with texture from ColorMap, TexMap, MVSTex, AdvTex-C, and ours respectively.
        \textbf{1) Compared to MVSTex:}
        Besides our method producing more complete geometry, we observe
        1.1) Scene 6: MVSTex produces an apparent split of the color on the carpet;
        1.2) Scene 7: MVSTex produces purple-like color at the far-end of the wall;
        1.3) Scene 8: for MVSTex, the top-middle of the image has an apparent color seam on the wall;
        1.4) Scene 9: MVSTex produces an apparent color seam at the far-end of the red ceramic ground;
        1.5) Scene 10: MVSTex produces apparent color seams on the door;
        1.6) Scene 11: MVSTex generates large color seams on the meadow.
        \textbf{2) Compared to AdvTex-C:}
        2.1) Scene 6: the carpet's pattern of our method is sharper and the color is smoother;
        2.2) Scene 7: our method generates a sharper pattern of the wall;
        2.3) Scene 8: our method produces a sharper boundary between the lower gray and upper white areas of the wall;
        2.4) Scene 9: our method produces sharper boundaries of the architecture and there are no color breaks on the ground;
        2.5) Scene 10: our reconstruction has a sharper boundary for the door while AdvTex-C fused the door and wall;
        2.6) Scene 11: our method generates sharper cracks on the wall.
        }
        \label{supp fig: qualitative highlights}
    \end{subfigure}%
    \vspace{-0.3cm}
    \caption{
      \textbf{Qualitative results on UofI Texture Scenes.}
      For each method, we show results for Scene 6 to 11 from left to right.
      Best viewed in color and zoomed-in.
    }
    \vspace{-0.3cm}
    \label{supp fig: qualitative}
\end{figure*}

We display qualitative results for the six scenes that are not shown in the main submission in~\figref{supp fig: qualitative}. Compared to AdvTex-C, our \texttt{TexInit} + \texttt{TexSmooth} framework largely reduces artifacts (\figref{supp fig: qualitative AdvTex}~\vs~\figref{supp fig: qualitative ours}), yielding more perceptual similarity while maintaining more sharpness (\figref{supp fig: qualitative highlights}).

In addition, we provide an HTML page in the supplementary to display more rendering comparisons from those 10\% evaluation views.

\end{document}